\documentclass[iicol,sn-apa]{sn-jnl}

\usepackage{graphicx}%
\usepackage{multirow}%
\usepackage{amsmath,amssymb,amsfonts}%
\usepackage{amsthm}%
\usepackage{mathrsfs}%
\usepackage[title]{appendix}%
\usepackage{xcolor}%
\usepackage{textcomp}%
\usepackage{manyfoot}%
\usepackage{booktabs}%
\usepackage{algorithm}%
\usepackage{algorithmicx}%
\usepackage{algpseudocode}%
\usepackage{listings}%
\usepackage{siunitx}

\theoremstyle{thmstyleone}%
\theoremstyle{thmstyletwo}%

\theoremstyle{thmstylethree}%

\begin{document}

\begingroup
\thispagestyle{empty}
\begin{minipage}{\textwidth}
    \large
    This version of the article has been accepted for publication in the \textit{Journal of Intelligent Manufacturing}. The Version of Record is available online at:~\url{https://doi.org/10.1007/s10845-025-02778-z}
\end{minipage}
\clearpage
\endgroup

\setcounter{page}{1}

\title[Article Title]{ISP-AD: A Large-Scale Real-World Dataset for Advancing Industrial Anomaly Detection with Synthetic and Real Defects}


\author*[1,2]{\fnm{Paul Josef} \sur{Krassnig}}\email{paul.krassnig@pccl.at}\email{paul-josef.krassnig@stud.unileoben.ac.at}

\author[1,2]{\fnm{Dieter Paul} \sur{Gruber}}\email{dieter.gruber@pccl.at}

\affil[1]{\orgname{Polymer Competence Center Leoben GmbH},
    Leoben, \country{Austria}}

\affil[2]{Chair of Materials Science and Testing of Polymers, \orgname{Montanuniversität Leoben},
    Leoben, \country{Austria}}


\abstract{Automatic visual inspection using machine learning plays a key role in achieving zero-defect policies in industry. Research on anomaly detection is constrained by the availability of datasets that capture complex defect appearances and imperfect imaging conditions, which are typical of production processes. Recent benchmarks indicate that most publicly available datasets are biased towards optimal imaging conditions, leading to an overestimation of their applicability in real-world industrial scenarios. To address this gap, we introduce the Industrial Screen Printing Anomaly Detection Dataset (ISP-AD). It presents challenging small and weakly contrasted surface defects embedded within structured patterns exhibiting high permitted design variability. To the best of our knowledge, it is the largest publicly available industrial dataset to date, including both synthetic and real defects collected directly from the factory floor. Beyond benchmarking recent unsupervised anomaly detection methods, experiments on a mixed supervised training strategy, incorporating both synthetic and real defects, were conducted. Experiments show that even a small amount of injected, weakly labeled real defects improves generalization. Furthermore, starting from training on purely synthetic defects, emerging real defective samples can be efficiently integrated into subsequent scalable training. Overall, model-free synthetic defects can provide a cold-start baseline, whereas a small number of injected real defects refine the decision boundary for previously unseen defect characteristics, thereby meeting key industrial inspection requirements such as low false positive rates and high recall. The presented unsupervised and supervised dataset splits are designed to emphasize research on unsupervised, self-supervised, and supervised approaches, enhancing their applicability to industrial settings. The dataset is publicly available at \url{https://doi.org/10.5281/zenodo.14911042}.}




\maketitle

\section{Introduction}\label{sec1}
With the affordability of modern computing power, the research and subsequent application of deep learning-based surface defect detection in industry is on the rise~\cite{Prunella.2023,Jha.2023}. As part of smart manufacturing and emerging Industry 5.0, recent publications address surface defect detection as an anomaly detection problem~\cite{Liu.2024,Lohweg.2023,Wen.2024}. In the context of surface defect inspection, an anomaly can be defined as any unwanted deviation from the sample's permitted surface variability and appearance, which could have known and unknown characteristics. Furthermore, the rare occurrence of anomalies compared to fault-free instances leads to heavily imbalanced data distributions~\cite{Bai.2024}. These anomalies can affect both the aesthetic and functional properties of the product and are therefore of major interest to industrial quality control.

Up to now manual inspection by humans is still part of industrial quality control. This repetitive task is prone to subjective assessment and fatigue resulting in quality fluctuations~\cite{chin.1982,Kujawinska.2015}. Integrating automated visual inspection on the factory floor improves product quality and efficiency by reducing human labor, subjective assessment, and subsequent production costs. Furthermore, its application is not limited to specific industries and is already used in the e.g. automotive, textile, electronics and agriculture industries~\cite{RaisulIslam.2024}.

A key element of any machine vision system is its underlying defect detection algorithm, enabling the decision-making process of categorizing a sample as fault-free (normal) or defective (anomalous)~\cite{Ren.2022}. The performance of machine-learning and more specific deep learning-based methods is dependent on the quality and amount of available data. Supervised algorithms based on deep convolutional neural networks (DCNN)~\cite{LeCun.2015} have gained remarkable performance in various defect detection tasks, relying on large labeled datasets~\cite{Zheng.2021,Saberironaghi.2023}.

However, collecting large amounts of fault-free and defective data and subsequent labeling is labor-intensive and often impractical in many industrial scenarios. Additionally, supervised methods struggle to generalize to unseen defects that were not part of the training data.
To address these issues, current research focuses on unsupervised anomaly detection methods~\cite{Cui.2023}. These methods solely rely on fault-free samples during training, learning feature representations of its underlying normal data distribution. With the publication of datasets such as MVTec~\cite{PaulBergmann.}, a wide range of different approaches have emerged, usually based on the comparison of image features or the reconstruction of normal image instances.

Extracted features, obtained by means of layers of e.g. DCNN, are applied in memory bank~\cite{Lee.2022b,Roth.2022,Xie.2023}, normalizing flow~\cite{Zhou.2024,Yu.2021,Tailanian.2024} or knowledge distillation-based approaches~\cite{Bergmann.2020,Rudolph.2023,Batzner.2024}. During inference, extracted test features are compared to learned normal representations using distance metrics or distribution mappings.

Reconstruction-based methods using autoencoders~\cite{Bergmann.2019} or generative adversarial networks (GAN)~\cite{Zhang.2022} attempt to reconstruct normal image regions while failing in resembling anomalies, resulting in anomaly scores.

These approaches suffer from poor reconstruction performance on fine-grained structures as well as demanding model training (mode collapse in GANs). Recently, diffusion-based approaches~\cite{Tebbe.2024,Mousakhan.25.05.2023,Zhang.15.03.2023} have gained increased attention, to address reconstruction limitations. These methods utilize iterative noising and denoising processes to model underlying data distributions at a computationally intensive cost.

Additional approaches are based on the synthesis of defects in both image and/or feature spaces, showing improved detection performance on image and pixel-level tasks~\cite{Liu.2024,Zavrtanik.02.08.2022}.

Due to the inherent nature of manufacturing processes, defective samples accumulate over time. Despite the promising detection performance of unsupervised methods, their ability to utilize defective data is limited. Consequently, incorporating both real and synthesized defects into the training process is emerging as a viable strategy, enhancing the method's discriminative capabilities.

In addition to defect detection performance, the industrial applicability of a defect detection method depends on meeting process requirements such as process cycle times, adaptability to different products, and robustness to permitted design variability and varying operational conditions. To evaluate model performance in an industrial setting, a dataset that reflects these conditions is crucial. However, a review of the literature shows that most publicly available datasets are generated under “laboratory conditions”, failing to capture the complexities of industrial environments~\cite{Liu.2024,Alzarooni.2025}. As a result, the benchmarks of state-of-the-art (SOTA) anomaly detection methods on these datasets are often overestimated compared to real-world industrial scenarios.

\subsection{Contributions}\label{sec11}

The aim of this publication is twofold. First, to bridge the gap to industry, we introduce the Industrial Screen Printing Anomaly Detection Dataset (ISP-AD), derived from a real-world industrial use case in screen printing. This dataset provides both unsupervised and supervised training data, comprising synthetic and real defects collected during production. Second, a mixed supervised training strategy is investigated to efficiently utilize the available weakly labeled data consisting of both synthetic and real defects.

We envision the ISP-AD dataset as a research and development resource for advancing anomaly detection methods under realistic industrial conditions. Furthermore, we hope that it will contribute to a more application-oriented perspective, by helping to define paradigms and approaches that align with real-world requirements.

Thus, the main contributions of this work can be summarized as follows:

\begin{itemize}
    \item \textbf{Introduction of the Industrial Screen Printing Anomaly Detection Dataset (ISP-AD):} a novel large-scale dataset of structured patterns captured using three different optical modalities. The dataset originates from a real-world industrial screen printing process and includes permitted process-specific variability. With a total of \num{312674} fault-free samples and \num{246375} defective samples (patches, of which \num{245664} are synthetic and 711 are real), it is assumed to be the largest publicly available industrial defect detection dataset to date, enabling both unsupervised and supervised training scenarios~(\nameref{data}).

    \item \textbf{Benchmarking of SOTA unsupervised methods:} comprehensive image- and pixel-level evaluations on ISP-AD highlight the challenges posed by small and weakly contrasted defects embedded within high permitted design variability, establishing ISP-AD as a demanding benchmark under industrial conditions.

    \item \textbf{Extended experimental investigation and formalization of mixed supervised training:} building on our earlier work on mixed training~\cite{HaselmannKrassnig.2022}, the incremental incorporation of weakly labeled real defects into synthetic cold-start training was analyzed across multiple modalities and defect fractions. The strategy was formalized as a \emph{stochastic batch-level injection scheme}, in which real defects replace patches in a balanced synthetic stream with an injection probability $p_{\text{inj}} \in [1/B, 0.25]$. Synthetic defects define an initial decision boundary, and real defects are incorporated in subsequent retraining stages for iterative refinement within the known process domain. Experimental results indicate that even small amounts of previously unseen defects can substantially enhance generalization, efficiently complementing synthetic training data. Ablations across ResNet~\cite{He.2016}, EfficientNet~\cite{Eff.2019}, and ConvNeXt~\cite{Liu_2022_CVPR} backbones confirm the architecture-agnostic scalability of the approach, demonstrating robustness under different accuracy–efficiency trade-offs and suitability for industrial deployment.

\end{itemize}

The structure of the research work is as follows: Section~\ref{sec2} gives an overview of emerging industrial anomaly detection methods as well as available datasets, and highlights existing research gaps. Section~\ref{sec3} introduces the real-world ISP-AD dataset, including the dataset generation process, available data splits, and its limitations. Section~\ref{sec4} describes the applied defect detection methods, the proposed mixed supervised training, and the investigated SOTA unsupervised methods. Section~\ref{sec5} presents the benchmark results across all three optical modalities. Finally, Section~\ref{sec6} concludes the work and outlines future research directions.

\section{State of the Art}\label{sec2}
\subsection{Supervision in Industrial Anomaly Detection}\label{sec21}

Unsupervised methods do not rely on defective samples during training, avoiding the risk of bias towards seen anomalies that can occur in supervised settings. However, the absence of knowledge about anomalous data results in a lack of discriminative features, making it challenging to distinguish subtle anomalies from normal data~\cite{SemiPush.c,CatchingSwans.}. This can result in false positives for normal samples with high permitted variability or overlooked defective areas.

To address this, augmentation methods have emerged that synthesize defective samples, enabling models to learn more discriminative features by incorporating these synthetic representations during self-supervised training tasks. Methods like CutPaste or NSA~\cite{Schluter.30.09.2021,Li.08.04.2021} synthesize defective samples by cropping patches from normal samples, augmenting them (e.g., resizing or rotating), and either pasting them or blending them into random positions using techniques like Poisson image editing~\cite{Perez.2023}.

In contrast to sampling from the examined data distribution, DRAEM~\cite{Zavrtanik.17.08.2021} generates diverse defect shapes by extracting textures from different domains (out-of-distribution)~\cite{Cimpoi.2014} using masks generated via Perlin noise~\cite{Perlin.1985}. Additional approaches use Gaussian or simplex noise~\cite{Perlin.2002} added onto normal images to generate synthetic anomalies~\cite{TranDinhTien.,CollDis.}. More realistic defects, referred to as in-distribution defects, can be achieved by generating defect textures on normal samples by means of random walks~\cite{Haselmann.2017}.

Recent trends leverage generative models such as GANs and diffusion-based approaches for defect synthesis~\cite{Gui.2025,Zhong.2023,Duan.2023,He.2023,Hu.2023}
GAN-based approaches rely on sufficient training data, including defective samples, and often struggle to synthesize fine grained patterns.

In contrast, denoising diffusion models aim to address these issues by creating more realistic defects while requiring fewer defective samples.

Besides defect synthesis in the image domain, recent methods have shown promising results in generating defects directly in the feature space, e.g., by introducing Gaussian noise~\cite{Liu.2023} or sampling from a set of codebook features~\cite{Zavrtanik.02.08.2022}. Global and Local Anomaly co-Synthesis Strategy (GLASS)~\cite{Chen.2024} combines constrained defect synthesis at both feature and image levels, resulting in ``near-in'' and ``far-from'' normal sample distribution anomalies, achieving SOTA performance on MVTec~\cite{PaulBergmann.}.

However, common problems of image-level synthesis are the lack of realism and diversity, while feature-level synthesis is hard to control but more efficient. As a consequence, additionally leveraging real defective samples accumulated during production is a complementary approach to increase feature diversity.

Research using both synthetic and real samples during training has been reported in binary classification and object detection~\cite{Posilovic.2021,PierreGutierrez.2021,HaselmannKrassnig.2022,Dey.2024} as well as in recent supervised anomaly detection approaches~\cite{Rolih.2024,HuiZhang.,CatchingSwans.}.

For instance,~\cite{Posilovic.2021} generated synthetic defects using a GAN and trained an object detection architecture with both real and synthetic data, achieving a 5~\% improvement in average precision compared to using only real defects. Similarly,~\cite{Dey.2024} reported enhanced defect detection performance on a small-sized dataset by additionally incorporating synthesized defects generated via a denoising diffusion approach. FuseDecode~\cite{Kozamernik.2025} proposes a novel autoencoder-based anomaly detection model initially trained on noisy, unlabeled data. The predictions assist in generating weakly labeled datasets, enabling mixed supervision with synthesized and collected real defects, thereby reducing labeling efforts.

Compared to supervised classification approaches, supervised anomaly detection methods reduce bias toward seen real defects by incorporating strategies such as out-of-distribution defect synthesis.

\subsection{Industrial Anomaly Detection Datasets}\label{sec22}

In addition to efforts in dataset preparation steps, including image capturing, data cleaning, preprocessing, and labeling, obstacles such as compliance standards for the investigated product impede dataset publication. As a result, many publicly available datasets are generated in laboratory settings, attempting to mimic industrial use cases.

Furthermore, due to the rare occurrence of defective samples, the artificial generation of defects is emphasized. This synthesis can be conducted at the image-level using algorithms (e.g., generative models or 3D rendering~\cite{Denninger.2019}) or manually on physical samples using appropriate tools.

To mitigate the need for defective samples during training, many industrial datasets are designed for unsupervised settings. In such a setting, defective data is only included in validation and/or test splits, minimizing efforts in defective data preparation.

\begin{table*}[htbp]
    \centering
    \caption{Comparison of SOTA anomaly detection datasets by sample size, defective data availability, and generation method (manual, synthetic, real-world). The proposed dataset, ISP-AD, introduces supervised and unsupervised data splits with synthetic and real defects, collected at the factory floor. In addition, ISP-AD addresses key challenges of industrial inspection that are underrepresented in existing datasets: (i) real-world imperfect imaging conditions and high permitted sample variations; (ii) imbalanced test data splits that reflect realistic inspection scenarios; and (iii) small and weakly contrasted defects on structured patterns relative to large sample areas.}
    \resizebox{\textwidth}{!}{
        \begin{tabular}{ccccccccc}
            \toprule
            \textbf{Dataset} & \textbf{\#Total}       & \textbf{\#Good}        & \textbf{\#Defects }          & \textbf{\#Defects Train} & \textbf{\#Classes} & \textbf{Modality} & \textbf{Defect Generation} \\
            \midrule
            MVTec            & \phantom{00}\num{5354} & \phantom{00}\num{4096} & \phantom{000}\num{1258}      & \phantom{000}0           & 15                 & rgb/mono          & manually                   \\
            VisA             & \phantom{0}\num{10821} & \phantom{00}\num{9621} & \phantom{000}\num{1200}      & \phantom{000}0           & 12                 & rgb               & manually                   \\
            Real-IAD         & \num{151050}           & \phantom{0}\num{99721} & \phantom{00}\num{51329}      & \phantom{000}0           & 30                 & rgb               & manually                   \\
            RAD              & \phantom{00}\num{1510} & \phantom{000}\num{286} & \phantom{000}1224            & \phantom{000}0           & \phantom{0}4       & rgb               & manually                   \\
            DAGM             & \phantom{0}\num{16100} & \phantom{0}\num{14000} & \phantom{000}2100            & 1050                     & 10                 & mono              & synthetic                  \\
            MIAD             & \num{105000}           & \phantom{0}\num{87500} & \phantom{00}\num{17500}      & \phantom{000}0           & \phantom{0}7       & rgb               & synthetic                  \\
            KolektorSDD2     & \phantom{00}\num{3335} & \phantom{00}\num{2979} & \phantom{0000}356            & \phantom{0}246           & \phantom{0}1       & rgb               & real-world                 \\
            Textile          & \phantom{00}\num{2150} & \phantom{00}\num{1814} & \phantom{0000}336            & \phantom{000}0           & \phantom{0}6       & rgb               & real-world                 \\
            BTAD             & \phantom{00}\num{2830} & \phantom{00}\num{2250} & \phantom{0000}580            & \phantom{000}0           & \phantom{0}3       & rgb               & real-world                 \\
            VAD              & \phantom{00}\num{5000} & \phantom{00}\num{3000} & \phantom{000}2000            & \num{1000}               & \phantom{0}1       & rgb               & real-world                 \\
            \textbf{ISP-AD}  & \num{559049}           & \num{312674}           & \phantom{00}711/\num{245664} & \phantom{0}446           & \phantom{0}3       & rgb/mono          & real-world/synthetic       \\
            \bottomrule
        \end{tabular}}
    \label{tab1}
\end{table*}%

Table~\ref{tab1} visualizes common industrial anomaly detection datasets. The MVTec dataset~\cite{PaulBergmann.} has significantly impacted the field by introducing an unsupervised anomaly detection dataset consisting of 15 industrial categories, grouped into 5 texture and 10 object-based classes. The dataset contains a total of 5354 images, including 1258 defective samples with ground truth masks that enable pixel-level evaluation. Images were mostly captured under highly controlled illumination conditions, and defects were manually generated to produce realistic visual appearances.

The VisA dataset~\cite{Zou.2022} extended these efforts by including objects with complex structures, such as printed circuit boards, and multiple instances resulting in a dataset twice the size of MVTec. Considering additional industrial scenarios, Real-IAD~\cite{Wang.2024} introduced a multi-view dataset, covering 30 classes of a variety of materials, such as plastic, wood and ceramics. In addition, RAD~\cite{Cheng.2024}, included uneven illuminations and blurry collections to imitate varying real-world inspection conditions.

Despite advancements in dataset size and increased data variability, the data generation process (including defective samples) was still performed manually, leaving domain gaps compared to factory-floor conditions.

Both DAGM~\cite{DAGM.2007} and MIAD~\cite{Bao.2023} are based on synthetic defect generation. DAGM, published in 2007, contains 10 synthetically generated classes summing up to \num{14000} fault-free background textures and 2100 defectives, one for each background texture. In contrast, the more recent MIAD contains \num{105000} images of various outdoor industrial maintenance scenarios.  MIAD leverages 3D graphics software~\cite{Denninger.2019} to render realistic 3D scenes with varying surface textures, backgrounds and viewpoints on both fault-free and defective objects.

In contrast to above described industrial datasets, including artificial generated defects, samples collected at industrial production lines are of particular interest. 
The KolektorSDD2~\cite{Bozic.2021} dataset addresses a practical real-world example, containing complex textured background structures. Defects in this dataset vary in size and shape, ranging from small scratches to large surface imperfections.

Another example is a textile dataset~\cite{Thomine.2024}, further referred to as Textile, that reflects imperfect industrial conditions, such as image blurring and environmental contamination. A further inspection application of three different industrial products, showcasing body and surface defects, is presented within the BTAD dataset~\cite{Mishra.2021}.

The recently published VAD dataset~\cite{Baitieva.2024} bridges the gap to supervised anomaly detection by additionally introducing 1000 defective training images, thereby extending the unsupervised setting. The investigated piezoelectric element in VAD is prone to structural defects (e.g., cracks or pollutions) as well as to logical defects (e.g. wire or solder position).

\section{Industrial Screen Printing Anomaly Detection Dataset: ISP-AD}\label{sec3}

Benchmarks on anomaly detection datasets such as MVTec~\cite{PaulBergmann.} already show high detection performance above 99~\% image-level Area Under the Receiver Operating Characteristic curve (AUROC), indicating less demanding inspection scenarios for SOTA anomaly detection methods.

Therefore, recent efforts in dataset generation focus on mimicking real-world inspection conditions by altering viewpoints, illumination, background, or product placement. However, a review of the literature still reveals a gap in the availability of large-scale datasets (several 10 thousand instances) captured at real-world production lines. Compared to clearly pronounced defect classes in certain object categories~\cite{PaulBergmann.,Mishra.2021,Zou.2022}, introducing small-scale and subtle defects is important for creating more challenging datasets. Additionally, most datasets are prepared for unsupervised settings, not considering additional defective training samples for supervised approaches.

The proposed industrial dataset aims to address known limitations by introducing the following features:

\begin{enumerate}[1.]
    \item Data captured from a real-world industrial manufacturing scenario.
    \item Small and weakly contrasted defects on structured patterns, relative to large sample areas.
    \item Three imaging modalities to enhance data variety.
    \item Large-scale industrial data splits applicable to both unsupervised and supervised settings. Test splits are imbalanced, typical to manufacturing data distributions (zero-defect policies).
    \item Synthetic and collected real defective data accumulated during production, suitable for additional supervision.
    \item High permitted sample variations arising from the sample itself, preprocessing, and imperfect imaging conditions.
\end{enumerate}

This section provides an overview of the sample under investigation, covering its defect classes, dataset generation process, specifications, and limitations.

\subsection{Industrial Screen Printing and Defect Classes}\label{sec31}

\begin{figure*}[h]
    \centering
    \includegraphics[width=0.80\textwidth]{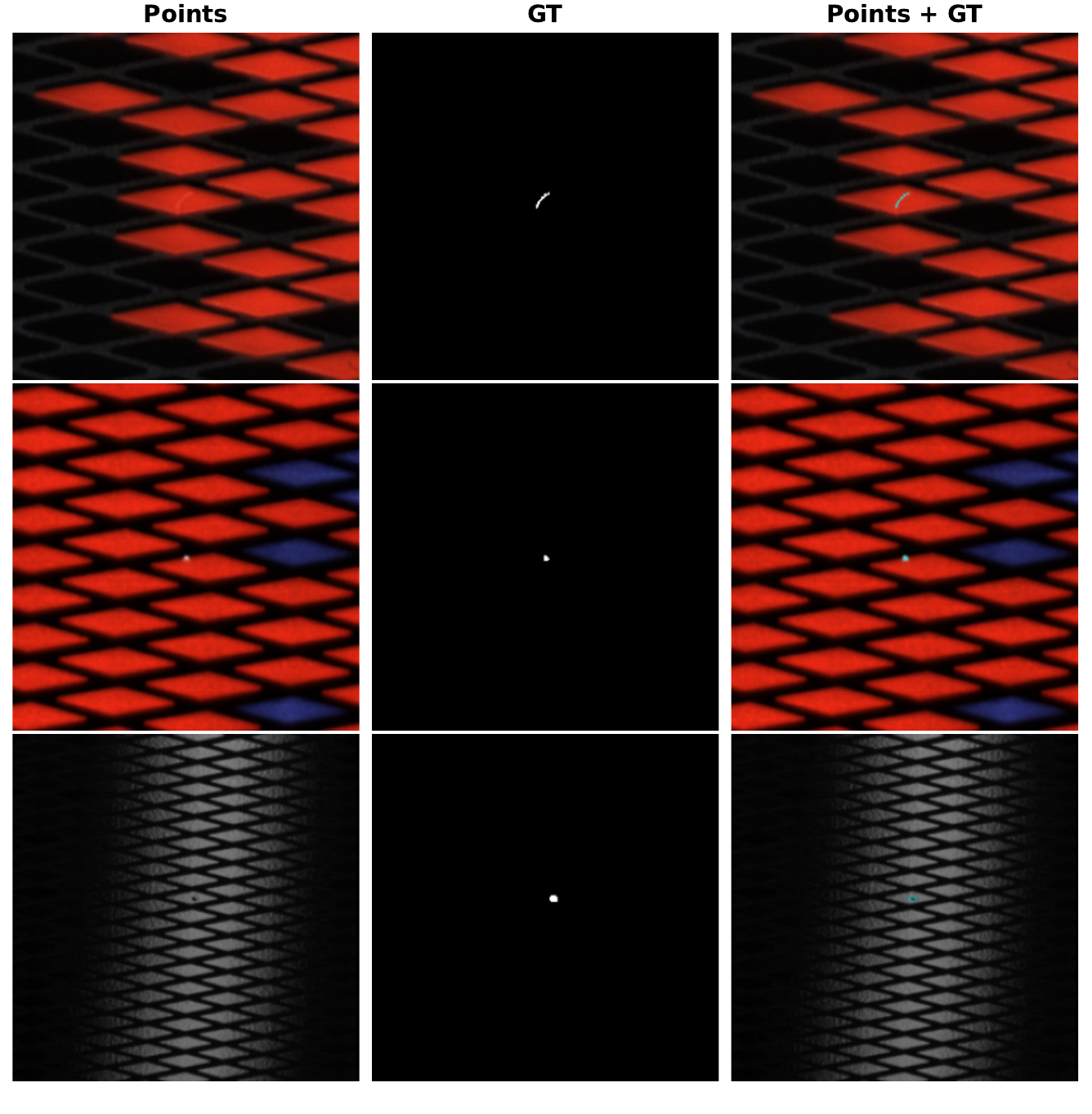}
    \caption{Examples of punctual defects from the ``Points'' defect group, their pixel-level ground truth masks (GT), and overlays (Points + GT). Each row corresponds to a certain defect (fiber, pinhole, mechanical deformation) within its imaging modality (top to bottom: LSM-1, LSM-2, ASM). The pinhole defect in row 2 appears small (few pixels in extension), whereas defects like the fiber in row 1 exhibit low contrast within high sample variability.}
    \label{fig1}
\end{figure*}

\begin{figure*}
    \centering
    \includegraphics[width=0.80\textwidth]{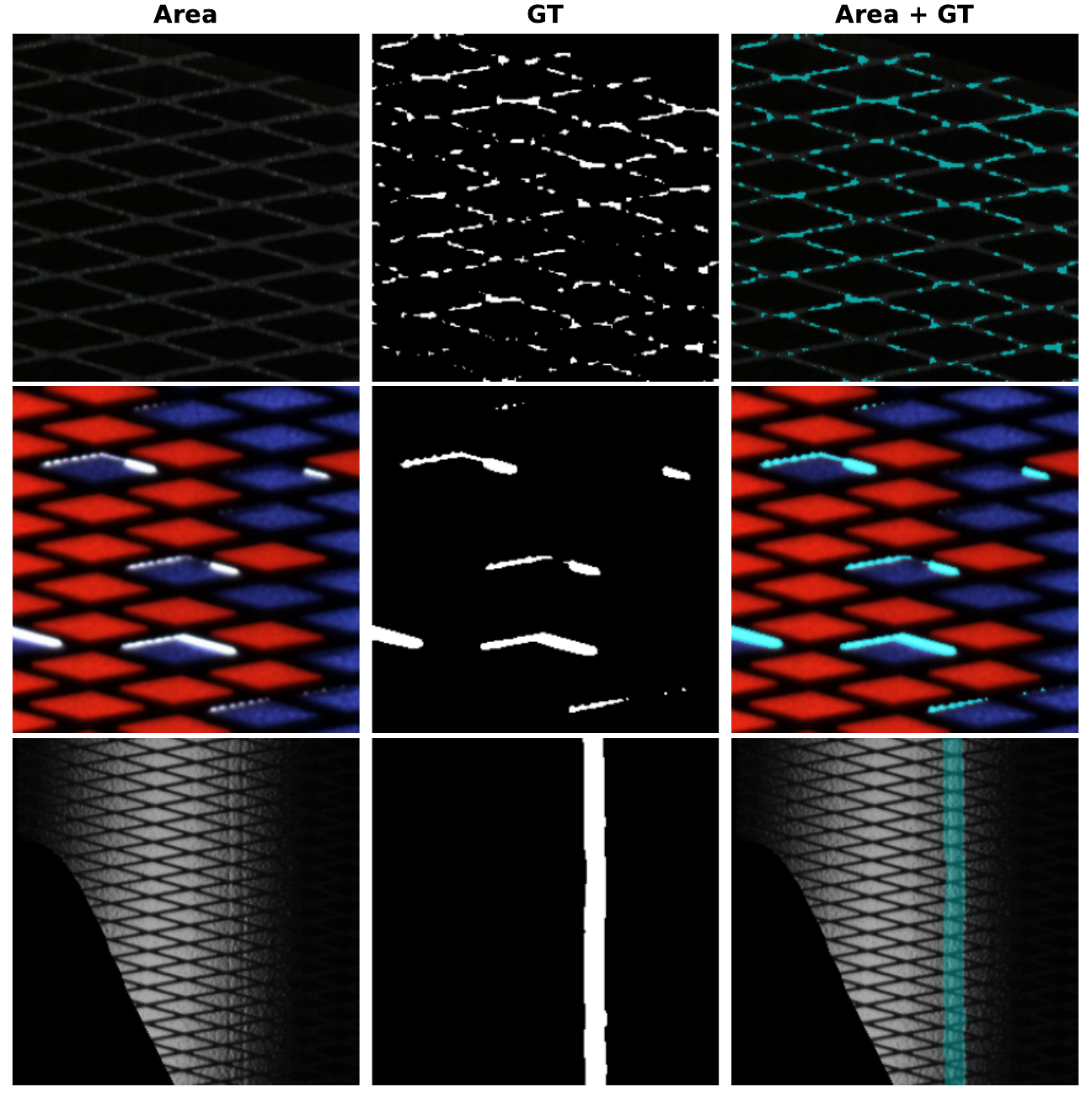}
    \caption{Examples of defects from the ``Area'' defect group, their pixel-level ground truth masks (GT), and overlays (Area + GT). Each row corresponds to a defect (grid, pattern misalignment, squeegee stroke) within its imaging modality (top to bottom: LSM-1, LSM-2, ASM). The pattern misalignment and grid defects exhibit distinct features across the entire patch, while the squeegee stroke appears within the transition area of direct reflection (bottom row).}
    \label{fig2}
\end{figure*}

The examined product is manufactured using a technique called screen or silk-screen printing~\cite{Biegeleisen.2012}. It is a low-cost and highly automatable manufacturing process applied in various industries e.g. automotive, textile and electronics~\cite{Sauer.2011}. In a nutshell, during the printing process, ink is deposited through a stencil with a predefined design onto the front and/or backside of a polymer carrier foil. The subsequent repetition using different stencils and colors results in a multilayer decorated foil plate with a high-quality appealing design. However, the complex manufacturing process is error-prone at basically every step of production. This results in a wide range of different defect appearances and classes. The investigated defect classes within this publication are visualized in Fig.~\ref{fig1} and Fig.~ \ref{fig2}.

They can be divided in punctual defects (Fig.~\ref{fig1}), such as inclusions, scratches, dots, pinholes, printing or screen defects or area defects (Fig.~\ref{fig2}), such as pattern misalignment, squeegee strokes and grid defects. As described in~\cite{KrassnigHaselmann.2022}, three different optical modalities, consisting of darkfield illumination in Line Scan Modality 1 (LSM-1), brightfield illumination in Area Scan Modality (ASM), and transmission illumination in Line Scan Modality 2 (LSM-2), were mandatory to visualize the wide range of different defect classes.

For example, most defects in the transparent top layer, such as scratches or mechanical deformations, are only visible under brightfield illumination used in ASM, whereas printing or screen defects are captured under darkfield conditions by means of LSM-1. In LSM-2, transmission illumination is used to detect defects such as pinholes or inhomogeneities within the print layers. For a detailed description of the imaging modalities and a visualization of different defect classes and their root causes, we refer to Sections 2 and 3 in~\cite{Krassnig.2024}.

In general, defects can be characterized as small (few px in extension in relation to large sized field of views (FOV) of up to \SI{e6}{\square\milli\metre}, with minimum object pixel sizes of approximately 75 µm) and weakly contrasted. They are embedded within structured background patterns that exhibit high permissible variability, making visual separation difficult due to low contrast, shown by the fiber in Fig.~\ref{fig1}. As visualized in Fig.~\ref{fig1} and Fig.~\ref{fig2}, defect characteristics are strongly dependent on both the imaging modality and the specific defect class. Furthermore, the intended transition regions of direct reflection, as depicted in ASM (see Fig.~\ref{fig_asm_artifacts}), introduce additional variations in contrast and intensity.

\subsection{Dataset Generation}\label{sec32}

The generation of appropriate datasets builds the basis for achieving the required defect detection performance of machine learning-based methods applied on the production line. Therefore, the design of validation and test splits mimicking the inspection process with its imbalanced data distributions is of utmost importance to perform reliable evaluations. Furthermore, generated data distributions must resemble the imperfect imaging conditions caused by environmental influences (e.g., varying illumination), data processing (varying background, viewpoint and segmentation artifacts), as well as permitted sample variations.

The proposed dataset is based on the work presented in~\cite{Krassnig.2024}, including its efficient data preprocessing and labeling workflow. All samples were collected directly from a real-world screen printing production line at a single industrial site, representing one product design. The investigated product is used in the automotive industry as a decorative element.

Prior to image acquisition, fault-free and defective samples were pre-sorted by a domain expert (quality engineer) in accordance with internal quality standards. This step ensured the exclusion of misplaced defective parts and borderline cases within the range of permitted normal variation. The sample state is characterized as fixed, thus additional temporal sample alterations can be excluded. Moreover, the samples were selected from different production batches, resembling the permitted design variability. Images were acquired by means of all three optical modalities (LSM-1, LSM-2, ASM) of the inspection system demonstrator installed on the factory floor. The acquired images of fault-free and defective samples were assigned to separate training and test splits. This step avoids any unwanted correlation of fault-free and defective patches during the subsequent patch extraction process. A detailed description of the utilized inspection system demonstrator, including its data processing steps, is provided in Section 4 of~\cite{Krassnig.2024}.

The extraction of augmented fault-free training patches (256 x 256 px) at random positions within the region of interest (ROI) is integrated into an automated procedure (steps 1 to 5 in Section 5.1~\cite{Krassnig.2024}).

Furthermore, following the central cropping in step 4, synthetic defects according to Section 5.2~\cite{Krassnig.2024} are generated in 50~\% of the fault-free patches.
The utilized algorithm, proposed by~\cite{Haselmann.2017}, consists of four steps:

\begin{enumerate}
    \item Generation of a binary defect skeleton, based on a stochastic process resembling a random walk with momentum.
    \item Generation of a random defect texture.
    \item Modification of the fault-free patch using the generated texture.
    \item Final assessment of defect visibility and rejection of synthesized defects below the visibility threshold.
\end{enumerate}

\begin{figure*}[h]
    \centering
    \includegraphics[width=0.80\textwidth]{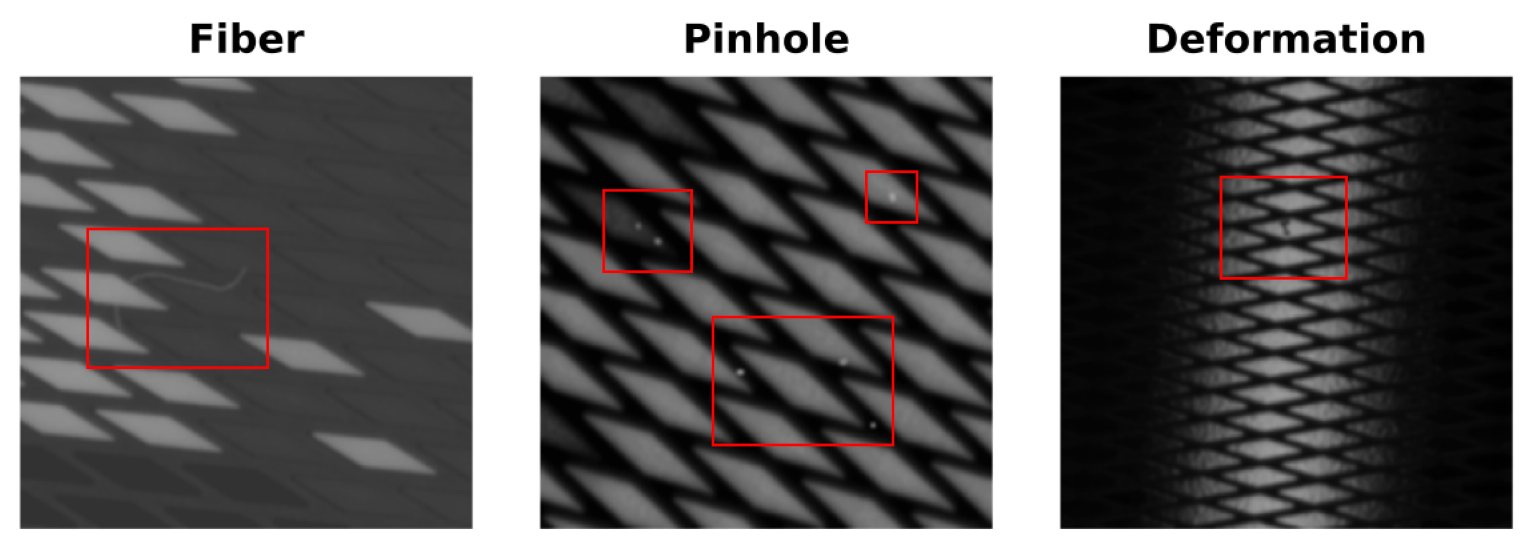}
    \caption{Synthesized defects generated using the algorithm proposed in~\cite{Haselmann.2017}. Hyperparameters were adjusted to simulate real punctual defects (e.g., pinholes in the middle) and elongated defects (e.g., bright contrasted fiber on the left; small mechanical deformation on the right patch) across all imaging modalities (left to right: LSM-1, LSM-2, ASM).}
    \label{fig3}
\end{figure*}

As described in Section 5.2 in~\cite{Krassnig.2024}, hyperparameters of the random variables were manually adjusted to synthesize bright- and dark-contrasted, punctuate and filamentous morphologies, aiming to imitate real-world defect characteristics, in all three optical modalities (Fig.~\ref{fig3}). As a result, balanced supervised training datasets -- also referred to as balanced synthetic streams -- were obtained.

Sufficient defect visibility was assessed using the sum of squared residuals of patch differences~\cite{Haselmann.2017}. A domain expert inspected up to 100 synthesized defects per split to tune defect-specific thresholds and hyperparameters. The resulting settings were applied to automatically synthesize tens of thousands of patches. Due to compliance regulations, a detailed listing of hyperparameters and their selected values was omitted. However, a full description of the algorithm is available in~\cite{Haselmann.2017}.

Fréchet Inception Distance (FID)~\cite{heusel2017gans} using Inception-v3~\cite{szegedy2016rethinking} (2048-dimensional features) was computed on \num{1000} synthetic point defects and on real point-defect groups from the train (\num{52}) and test (\num{46}) sets in LSM-1. All computations used unaugmented grayscale patches. Approximate alignment between the synthesized and real defect distributions was indicated: real-train vs.\ real-test resulted in an FID of approximately \num{54}, and (real-train or real-test) vs.\ synthetic-train in FIDs of approximately \numrange{54}{57} (see Limitations in Section~\ref{sec325}).

In addition, real defective patches were manually extracted on selected defective samples in the training and test splits. Thereby, central defect labeling with patch sizes of 512 x 512~px ensures proper defect positioning within augmented patches, which are cropped to a final size of 256 × 256~px during training.

Above described workflow minimizes the manual labeling effort to known defective samples and regions. The underlying image-level labels can be considered ``weak'' since no ground truth masks are provided, thereby eliminating the need for elaborate pixel-level annotation. In general, patch-based processing is an efficient augmentation technique in upscaling available data, enabling extraction of several \num{100000} patches from a few acquired samples with large FOVs. Theoretically, this automatic patch extraction process is unlimited, although it is restricted by available fault-free samples and FOV sizes~\cite{Krassnig.2024}.

Fault-free test data was generated using a sliding window with a stride of 160 px, extracting overlapping patches (256 x 256 px) to cover a grid. The same patch extraction process was applied to the training data for the unsupervised methods.

As an additional data cleaning step, unwanted defective patches (e.g., contamination in the process environment or defects overlooked by the domain expert on the factory floor) were removed from the unsupervised training and general test data.

As described earlier, the automatically extracted fault-free supervised training data was not further screened and can therefore be considered noisy. Based on a manual inspection of 500 randomly extracted and preprocessed (contrast enhanced) patches from LSM-1, it was estimated that 2 -- 5~\% contained overlooked defects, ranging from borderline cases to clearly contrasted defects such as dust and fibers.

To enable the evaluation of detection performance at the pixel-level, ground truth masks for the test patches were generated using the ilastik labeling tool~\cite{Berg.2019}, which is based on a random forest classifier. Following the assignment of pixel object classes, additional image processing techniques (e.g., erosion and dilation) were applied to ensure distinct defective regions. Examples of selected defective patches, ground truth masks (GT), and their overlays are shown in Fig.~\ref{fig1} and~\ref{fig2}.

A single domain expert (machine learning engineer) conducted the manual screening of the unsupervised training and general test data. The same expert produced patch-wise annotations for the weakly labeled defective training data and pixel-level ground-truth annotations for the test set to minimize inter-annotator bias.

\subsubsection{Data Splits}\label{sec321}

The aim of this publication is to advance research in solving real-world industrial inspection problems by leveraging both synthetic and available real defects accumulated during production processes. Therefore, supervised and unsupervised datasets for each optical modality, acquired from the factory floor, have been created. The supervised training splits include up to approximately \num{100000} augmented fault-free patches and \num{100000} synthetic defective patches, forming a large-scale balanced synthetic training stream. Additionally, real defective patches, extracted from defective samples representing various defect classes, are categorized into two groups: ``area'' and ``points''. For unsupervised approaches, the modality LSM-1 provided the largest number of fault-free samples, allowing extraction of up to 3678 manually screened training patches. Furthermore, reduced training splits are made available, each comprising up to 500 patches per modality. These smaller splits are designed to minimize computational effort for e.g. memory demanding methods.

Table~\ref{tabA1} summarizes the proposed dataset, including its synthetic and real defects, along with the distribution of defect groups (area, points) available for supervision.

\subsubsection{Data Format}\label{sec322}

In general, LSM-1 and LSM-2 patches are provided as 8-bit (rgb) .png files, while ASM patches are provided as single-channel 8-bit .png files. Ground truth masks are stored as single-channel 8-bit .png files. In the case of supervised training data, all patches are stored as single-channel 8-bit within the .hdf5 large file storage format. In addition to the image-level labels, pixel-level GTs are available due to defect synthesis. The folder structure of the unsupervised datasets follows the official MVTec~\cite{PaulBergmann.} scheme.

\subsubsection{Augmentation Settings}\label{sec323}

According to step 3 in Section 5.1~\cite{Krassnig.2024}, the following augmentation settings were applied to the supervised training data: vertical and horizontal flipping, rotation, shearing, and scaling. Random translation is not included in this list, as it is part of the prior random patch extraction process. Random affine transformations were applied to 80 -- 95~\% of the extracted patches. In addition, random contrast and brightness adjustments were applied. The corresponding transformation parameters with its values and ranges are shown in Table~\ref{tabA2}.

\subsubsection{Additional Data}\label{sec324}

In addition to proposed data splits, supplementary fault-free patches for all modalities are provided. These patches are the unaugmented counterparts to the fault-free supervised training data described above. They consist of \num{20000} and \num{10000} fault-free patches in rgb format for LSM-1 and LSM-2, and \num{20000} patches in single-channel format for ASM, all stored as .png files.

For further details, including a visual presentation of the dataset structure and available splits, we refer the reader to the README.md file in the dataset repository~(\nameref{data}).

\subsubsection{Limitations}\label{sec325}
Due to the limited amount of defective data available, a validation split was omitted. Although the fault-free patches in the test splits were extensively screened, it cannot be guaranteed that the labels are entirely defect-free.

In ASM, the manual extraction of defective patches introduced additional masked borders on the left and right patch sides (a few pixels in width), which could potentially lead to false positive predictions in non-robust unsupervised approaches, such as reconstruction-based methods. Therefore, cropping these regions during evaluation is recommended to mitigate this effect. Furthermore, the same defects may appear in different patch positions. However, due to varying illumination conditions in this modality, this can be considered an additional form of augmentation.

General limitations include the presentation of a single product design due to compliance regulations, as well as a constrained set of defect classes. However, the observed defect diversity arises from process-related variability and is not strictly limited by the design itself. In addition, ground truth masks for real defective training data are not included, as the focus was set on weakly labeled training data conditions. Although, their inclusion may be considered in future dataset revisions.

Estimated absolute FID scores may be inflated by permitted data variation, the small spatial extent of defects, and small sample sizes of real data (unsupervised train-good vs. test-good: approximately 38; see Table~\ref{tabA1} for data split details). Therefore, FID is interpreted relative to baselines, and defect-focused crops are recommended for future analyses. Since Inception features are ImageNet-biased, domain-tuned pretraining on ISP-AD is preferable to better capture fine-grained textures in future defect realism assessments. Given ongoing research on the human-perceptual visibility of surface defects~\cite{10.1117/12.3078056}, automatic defect evaluation remains challenging, and metrics such as FID have inherent limitations. Consequently, final expert assessment and evaluation of downstream performance are crucial.

ISP-AD is not intended to serve as a general-purpose anomaly detection dataset encompassing all possible defect classes encountered in manufacturing. While grounded in silk-screen printing, ISP-AD is designed as a benchmark for evaluating methods targeting similar industrial inspection challenges.

\section{Defect Detection Methods}\label{sec4}

The recently published work~\cite{Krassnig.2024} introduced an efficient data preprocessing workflow utilizing weakly labeled defective data in a supervised training approach. Furthermore, the possible extension to a mixed training strategy using both synthetic and available real defective data~\cite{HaselmannKrassnig.2022} was outlined, enhancing defect detection sensitivity of overlooked defect classes.

Therefore, the evaluation of the mixed training strategy on the presented demanding imaging modalities should give further insights in leveraging supervised methods in industrial anomaly detection. For comparison several SOTA unsupervised anomaly detection methods are investigated. Utilized approaches as well as their training settings will be described in the following section.
\subsection{Mixed Supervised Training}\label{sec41}

In industrial production, defects emerge at various stages of the manufacturing process and may be unavailable during training or even previously unknown.
Thus, defects can be categorized as:
\begin{itemize}
    \item \textbf{Seen defects}: Known defect classes that are available during training (synthetic or real).
    \item \textbf{Unseen defects}: Unknown or unavailable defect classes that may exhibit previously unobserved feature distributions.
\end{itemize}

The proposed mixed supervised training strategy showed increased detection performance by leveraging both synthetic and real defects during training, compared to training separately on synthetic or real ones~\cite{HaselmannKrassnig.2022}. Thus, depending on the availability of real defects, the method can be applied in cold-start scenarios, starting with purely synthetic training and transitioning to a mixed setup as real defects accumulate. This is an important capability in industrial applications, as the model can be improved step by step with newly available defective samples.

However, the ability to synthesize certain defect classes using the algorithm in~\cite{Haselmann.2017}, such as squeegee strokes shown in Fig.~\ref{fig2}, is limited or even infeasible. Nonetheless, such model-free approaches, as described in~\cite{wang2025stones}, do not rely on real defective samples for subsequent model training or conditioning, unlike generative models such as GAN- or diffusion-based methods~\cite{Duan.2023,Hu.2023,Gui.2025}. They may therefore represent a more deployment-friendly strategy, particularly in cold-start scenarios where no real defective samples are available. In other words, model-free anomaly synthesis refers to methods that generate artificial defects through handcrafted, augmentation-based, or procedural rules without training a generative model~\cite{Li.08.04.2021,Zavrtanik.17.08.2021,Schluter.30.09.2021,Haselmann.2017}.

Thus, even small amounts of previously unseen real defects may contribute to learning feature representations that could not be captured by synthesis or were absent in the available real data. Additionally, incorporating real defects, similar to synthesized ones, increases the feature diversity of already seen defects. Throughout this work, we use the term \emph{few-shot} to refer specifically to training or defect modeling regimes with only a limited amount of real defective samples available (e.g., 10 patches), while the majority of training data is still provided by synthetic defective or fault-free patches (see Section~\ref{sec321} on data splits).

Utilizing the above-introduced datasets, a DCNN pretrained on ImageNet~\cite{JiaDeng.2009} was trained in a supervised manner. Synthetic defective patches were generated in 50~\% of the fault-free patches.

During training, each patch in this balanced synthetic stream was replaced by an augmented real defective patch with an injection probability  $p_{\text{inj}}$, chosen within the range

\begin{equation}
    p_{\text{inj}} \in \left[\tfrac{1}{B},\,0.25\right],
\end{equation}
\vspace{0.5em}

and kept constant during each training stage, where $B$ denotes the batch size.

\vspace{1.0em}
The injection probability $p_{\text{inj}}$ can be tuned to the quantity and diversity of available real defects: lower values favor specificity, while higher values improve sensitivity but also increase the risk of bias toward injected feature distributions. Therefore, the lower bound ensures that, in expectation, at least one real defect is injected into each batch, while the upper bound of 25~\% was chosen empirically to prevent real defects from dominating the synthetic stream.

Unlike dataset-level mixing approaches~\cite{Posilovic.2021,Dey.2024} or deterministic batch-level schemes that enforce one real defect per batch~\cite{huang2025weakly}, our method introduces a tunable stochastic injection. This allows adaptation to defect availability and synthesis capabilities while maintaining variability across training batches.

The proposed \textit{stochastic batch-level injection scheme} treats each patch independently. Consequently, the expected fraction of real defects in any batch equals $p_{\text{inj}}$, regardless of whether the underlying stream consists of purely fault-free patches or of a balanced synthetic fault-free mixture.

This formulation highlights the connection to our previously published oversampling approach~\cite{Krassnig.2024}: when the balanced synthetic stream is replaced by a pure fault-free stream and the injection probability is set to $p_{\text{inj}} = 0.5$, the two approaches converge. In this special case, each batch contains, in expectation, 50~\% fault-free and 50~\% real defective patches.

Supervised binary classification is inherently a closed-set problem, in contrast to the open-set problem of anomaly detection described above. In production environments, such as the investigated screen printing process, possible defect causes are often known in advance, even if corresponding defective samples are not yet available. This reduces the defect space to a restricted open-set problem (see Appendix~\ref{ROS-AD}), where not all defects are yet observed, but their characteristics are likely to fall within a range of expected or process-related visual features. To address this setting, our mixed supervised training can be characterized as an adaptable closed-set strategy, operating within this bounded defect space defined by process-specific visual characteristics, both observed and yet unobserved.
Adapting the classifier to emerging, previously unseen defect classes with a minimal set of recently available samples would be an important capability in industrial settings.

The following experiments in Section~\ref{sec52} aim to investigate the generalization capabilities of the proposed mixed training strategy for previously unseen defects (e.g., defects that could not be synthesized). In addition, its scalability is examined under varying proportions of collected real defects across all imaging modalities: LSM-1, LSM-2, and ASM. Before the main evaluations, ablation experiments across various backbones were performed.

\subsection{Training Settings}\label{sec42}

The network training was performed using the stochastic gradient descent optimizer with the following parameters: a learning rate ranging from \SI{1e-5} to \SI{5e-4}, weight decay of \SI{1e-2}, and momentum of 0.9. Cosine annealing with warm restarts ($To=7813$, $T\_mult=2$) was applied as the learning rate scheduler, following the approach outlined in~\cite{Loshchilov.2016}.
Given a batch size of 128 and the dataset size (Section~\ref{sec321}), the first scheduler restart occurs after five epochs (\num{7813} iterations), with subsequent restart intervals doubling.

Based on a backbone ablation experiment comparing ResNet~\cite{He.2016}, EfficientNet~\cite{Eff.2019}, and ConvNeXt~\cite{Liu_2022_CVPR}, ResNet-18 pre-trained on ImageNet~\cite{JiaDeng.2009} was selected as the backbone, as it provided the best accuracy–efficiency trade-off for the underlying use case (see Section~\ref{sec520}). The final fully connected layer, originally consisting of 1000 output neurons (as per ImageNet pretraining), was replaced by a layer with two output neurons representing the binary class labels ``good'' and ``defective''. To mitigate potential domain bias from pretrained weights, all layers of the selected architecture were fine-tuned using the initial learning rate and scheduling scheme described above. As additional regularization, dropout with a rate of $10\%$ was applied to the final fully connected layer of the classifier head.

As part of the preprocessing, patches from LSM-1 were brightness-adjusted by a factor of 1.5. Additionally, weak smoothing was applied across all modalities using a Gaussian kernel with a kernel size of 3 and a sigma of 1. In mixed training scenarios using both synthetic and real defects, the injection probability $p_{\text{inj}}$ for real defects was set to $1/32$. Thus, for a batch size of 128, the expected number of augmented real defects per batch was four. This setting aims to prevent overrepresentation of real defects within the balanced synthetic data stream.

Mixed precision training was carried out until the validation loss showed no improvement for 20 consecutive epochs. Model selection was based on the lowest validation loss. As previously described, validation was performed on the available test set. For the binary classification task, thresholds were determined according to the optimal F1-scores obtained.

\subsection{Unsupervised Methods}\label{sec43}
In order to compare the defect detection performance to SOTA unsupervised approaches, methods for feature embedding, reconstruction and synthesis are investigated. Thus, the methods used are grouped according to their approaches and thus briefly described. In addition, method and training settings utilized in the experiments are listed.

\subsubsection{Normalizing Flows}\label{sec431}

In contrast to utilizing large memory banks for feature comparison, normalizing flows model normal feature distributions. During training, the initial complex normal feature distribution is transformed into a standard normal distribution through a series of invertible mappings. In contrast to DCNNs as feature extractors, U-Flow~\cite{Tailanian.2024} utilizes a multiscale vision transformer architecture~\cite{DosovitskiyAlexey.2020}, pretrained independently for each scale. A fully invertible architecture is achieved by adapting a UNet-like structure~\cite{Ronneberger.2015} to normalizing flows. Thus, the feature extractor acts as the encoder, while the normalizing flow serves as the decoder.

\subsubsection{Student-Teacher}\label{sec432}

Within this approach, a shallow student network attempts to mimic the output of a pre-trained (distilled) teacher using knowledge distillation. Exclusively trained on fault-free data, the distilled student network fails to predict the teacher's output on anomalous features, resulting in an anomaly score. EfficientAD~\cite{Batzner.2024} provides a lightweight network architecture with restricted receptive field to accelerate feature extraction. Additionally, a hard feature loss function and a loss penalty are introduced to prevent the student from generalizing its imitation to out-of-distribution images. Furthermore, an integrated autoencoder ensures detection of logical anomalies.

\subsubsection{Denoising Diffusion}\label{sec433}

Denoising diffusion probabilistic models are a class of generative models inspired by non-equilibrium thermodynamics. They consist of a forward diffusion process that gradually adds Gaussian noise to the input image over several steps, transforming it into a standard normal distribution. During training, the reverse diffusion process learns to denoise the corrupted image step by step by minimizing the difference (mean squared error) between the predicted noise and the actual noise added during the forward process. The method proposed in Denoising Diffusion Anomaly Detection (DDAD)~\cite{Mousakhan.25.05.2023} utilizes target images to guide the denoising process, improving its ability to reconstruct normal patterns. Anomaly localization is achieved through both feature-wise and pixel-wise comparisons between the reconstructed image and the input image. The pretrained feature extractor is adapted to the target domain using examples generated by the denoising model, further enhancing detection performance.

\subsubsection{Defect Synthesis}\label{sec434}

GLASS~\cite{Chen.2024} extends purely image-level anomaly synthesis by introducing Global Anomaly Synthesis (GAS) at the feature level and Local Anomaly Synthesis (LAS) at the image level. GAS utilizes Gaussian noise guided by gradient ascent and truncated projection to synthesize anomalies in the feature space near the normal sample distribution (in-distribution anomalies). LAS generates out-of-distribution anomalies by overlaying textures on normal images using Perlin masks, similar to the approach in~\cite{Zavrtanik.17.08.2021}. Finally, a segmentation network is trained end-to-end using three loss functions:  normal feature, local anomaly feature, and global anomaly feature loss. During inference, only the normal feature branch is utilized for anomaly detection.

\subsubsection{Implementation Details}\label{sec435}

To standardize the evaluation workflow, the Anomalib API~\cite{Akcay.2022} was utilized for the following unsupervised methods~\cite{Tailanian.2024,Batzner.2024}. Therefore, a training and evaluation pipeline, was implemented. Default method settings, as documented in the model configuration files, were applied. In case of EfficientAD~\cite{Batzner.2024}, the medium (M) patch descriptor network was chosen.

Additionally, the official implementations of DDAD~\cite{Mousakhan.25.05.2023} and GLASS~\cite{Chen.2024}, along with their configuration files, were selected. The target image conditioning parameters $w$ and $w\_DA$ were set to 3, together with 4 domain adaptation epochs, as specified in DDAD. In GLASS, the manifold hypothesis was selected as the GAS strategy, while foreground masks were omitted in LAS.

In general, the training input shape for image patches was set to 256 x 256 px. Despite center cropping (224 x 224 px) in DDAD for LSM-1, and center cropping (224 x 224 px) in ASM, no additional preprocessing was applied to the raw image patches. Center cropping within ASM excludes any masked borders produced by manual patch extraction (Section~\ref{sec325}). Due to EfficientAD's autoencoder architecture, the input shape was kept constant at 256 x 256 px, however during evaluation, border regions were omitted, as shown in Fig.~\ref{fig4}. To enable feature extraction by means of the multi-scale vision transformer backbone in U-Flow, patches were resized to 448 x 448 px.

Configuration files, including the selected hyperparameters and preprocessing settings for all methods, are publicly available in the dataset repository~(\nameref{data}).

\subsubsection{Training Settings}\label{sec436}

Anomalib's engine was trained until the pixel-level AUROC showed no improvement for 30 consecutive epochs, with a limit of 300 epochs. The denoising diffusion approach was trained for 3000 steps with learning rates ranging from \SI{5e-5} to \SI{1e-4}. During testing, the starting point for the denoising trajectory was set to 250, with a step size of 25. Due to memory limitations, the batch size for both training and testing was reduced to 8. GLASS was trained up to 160 meta-epochs, with learning rates ranging from \SI{1.25e-5} to \SI{5e-5}, also using a batch size of 8. Thresholds for both image- and pixel-level metrics were determined by computing the optimal F1-score at the image-level. Training was performed on reduced splits of up to 500 patches.
\subsubsection{Hardware Setup}\label{sec437}

Experiments were performed on desktop workstations, equipped with NVIDIA GPUs (GeForce RTX 3090 or GeForce RTX 4080), AMD 16-core processors and 64GB RAM running on OS Windows 10/11. The Python environments (version $\geq$ 3.9.15) utilized GPU versions of the PyTorch framework (version $\geq$ 2.1.0) with CUDA Toolkit version 11.8, as well as PyTorch Lightning (2.4.0).

\section{Experiments and Results}\label{sec5}

The following section investigates the defect detection performance of the supervised and unsupervised approaches proposed in Section~\ref{sec4}, applied across all optical modalities: LSM-1, LSM-2, and ASM.

\begin{table*}[t]
    \centering
    \caption{Ablation study comparing oversampling~\cite{Krassnig.2024} and mixed training ($p_{\text{inj}} = 1/32$)~\cite{HaselmannKrassnig.2022} across different backbones (ResNet~\cite{He.2016}, EfficientNet~\cite{Eff.2019}, ConvNeXt~\cite{Liu_2022_CVPR}), different fractions of real defects, and a synthetic-only baseline on LSM-1.
        Metrics: MCC, inference time per patch, and throughput (mean~$\pm$~std, patches/s).
        The best MCC in the \textbf{full-data regime} (172 real defects) is highlighted in \textbf{bold}, while the best MCC in the \emph{few-shot regime} (10 real defects) is marked with a dagger ($\dagger$).}
    \label{tab:ablation}
    \resizebox{0.95\textwidth}{!}{%
        \begin{tabular}{l l c
                S[table-format=1.2]
                S[table-format=1.2]
                S[table-format=4.0(2)]}
            \toprule
            \textbf{Model}  & \textbf{Strategy}       & \textbf{\#Real Defects}         &
            \textbf{MCC}    & \textbf{Inf. Time (ms)} & \textbf{Throughput (patches/s)}                                     \\
            \midrule
            ResNet-18       & Oversampling            & \phantom{0}10                   & 0.70             & 0.16 & 6309(5) \\
                            &                         & 172                             & 0.91             &      &         \\
                            & Mixed                   & \phantom{00}0                   & 0.81             &      &         \\
                            &                         & \phantom{0}10                   & 0.91$^{\dagger}$ &      &         \\
                            &                         & 172                             & \textbf{0.96}    &      &         \\
            \midrule
            ResNet-50       & Oversampling            & \phantom{0}10                   & 0.78             & 0.45 & 2216(1) \\
                            &                         & 172                             & 0.93             &      &         \\
                            & Mixed                   & \phantom{00}0                   & 0.74             &      &         \\
                            &                         & \phantom{0}10                   & 0.91$^{\dagger}$ &      &         \\
                            &                         & 172                             & \textbf{0.97}    &      &         \\
            \midrule
            EfficientNet-B0 & Oversampling            & \phantom{0}10                   & 0.73             & 0.22 & 4608(5) \\
                            &                         & 172                             & 0.94             &      &         \\
                            & Mixed                   & \phantom{00}0                   & 0.86             &      &         \\
                            &                         & \phantom{0}10                   & 0.91$^{\dagger}$ &      &         \\
                            &                         & 172                             & \textbf{0.96}    &      &         \\
            \midrule
            EfficientNet-B4 & Oversampling            & \phantom{0}10                   & 0.74             & 0.55 & 1815(1) \\
                            &                         & 172                             & \textbf{0.98}    &      &         \\
                            & Mixed                   & \phantom{00}0                   & 0.87             &      &         \\
                            &                         & \phantom{0}10                   & 0.94$^{\dagger}$ &      &         \\
                            &                         & 172                             & 0.97             &      &         \\
            \midrule
            ConvNeXt-Tiny   & Oversampling            & \phantom{0}10                   & 0.80             & 0.52 & 1912(9) \\
                            &                         & 172                             & \textbf{0.98}    &      &         \\
                            & Mixed                   & \phantom{00}0                   & 0.92             &      &         \\
                            &                         & \phantom{0}10                   & 0.96$^{\dagger}$ &      &         \\
                            &                         & 172                             & \textbf{0.98}    &      &         \\
            \bottomrule
        \end{tabular}}
\end{table*}

\subsection{Defect Detection Performance Metrics}\label{sec51}

In addition to commonly used metrics such as recall, false positive rate (FPR), and AUROC, metrics designed to handle class imbalance effectively were selected. For image-level metrics, the Matthews Correlation Coefficient (MCC) was applied. The MCC, which ranges from -1 (indicating inverse prediction) to 1 (indicating perfect prediction), incorporates all entries of the confusion matrix: true negatives (TN), true positives (TP), false negatives (FN), and false positives (FP). In this context, TN represents correct fault-free predictions, while TP represents correctly predicted defective patches. To account for pixel-level class imbalance, the per region overlap score (PRO) was chosen to assign appropriate weightings to different sized defective regions.

\subsection{Mixed Supervised Training}\label{sec52}

To evaluate the scalability of the introduced mixed supervised training, different fractions of real defects accumulated during production were injected into the balanced stream of fault-free and synthetically generated defects. Additionally, to examine adaptability to unseen features, such as those missed during synthesis, training was performed exclusively on each defect group (points or area).

Following random shuffling of all available defects, fractions of 1/2, 1/4, 1/8, and 1/16 were extracted in an ordered manner. This approach ensures that defects included in smaller fractions are also represented in larger ones. These fractions consist of both defect groups, points (Fig.~\ref{fig1} ) and area (Fig.~\ref{fig2}), representing features that are either punctual or distributed across large sample areas (mixed group). In all experiments, starting from purely synthetic training, real defects were injected with a probability $p_{\text{inj}}$ of 1/32. To ensure comparability between experiments, the same training settings, including learning rates and seeds, were applied within each optical modality.

Prior to investigating the scalability and generalizability of mixed supervised training, ablation experiments were conducted on LSM-1 to identify an industry-applicable backbone that satisfies process-cycle constraints defined in~\cite{Krassnig.2024}. The results, summarized in Table~\ref{tab:ablation}, compare the oversampling approach proposed in~\cite{Krassnig.2024} with mixed training~\cite{HaselmannKrassnig.2022} across different backbones, defect fractions, and synthetic-only baselines, thereby highlighting accuracy–efficiency trade-offs relevant for industrial inspection. Following the backbone selection, image-level defect detection performance in LSM-1 and ASM across various defect fractions is presented in Tables~\ref{tab2} and~\ref{tab3}.

\subsubsection{Ablation: Backbone Selection}\label{sec520}

As defined in~\cite{Krassnig.2024}, the overall inspection time, utilizing all three measurement chambers, was restricted to 15–30 s. To avoid bottlenecks induced by the inference stage, efficient backbones from standard ResNet architectures~\cite{He.2016} as well as more recent EfficientNet and ConvNeXt architectures~\cite{Eff.2019, Liu_2022_CVPR} were evaluated.

The inference time measurements reported in Table~\ref{tab:ablation} were determined as the arithmetic mean of five cycles with 100 repetitions, including preprocessing of the respective methods (GeForce RTX 4080 setup as described in Section~\ref{sec437}). Each cycle was preceded by a 10-iteration warm-up. To account for asynchronous CUDA processing, inference times were measured using PyTorch’s synchronized CUDA \textit{events} at a batch size of 128, with \textit{autocast} applied for mixed-precision evaluation.

As shown in Table~\ref{tab:ablation}, all architectures achieved fast inference times below 1 ms per patch, although the shallow ResNet-18 outperformed the other backbones by a factor of 1.36 compared to EfficientNet-B0 and up to approximately 3.5 compared to EfficientNet-B4 and ConvNeXt-Tiny.

ConvNeXt-Tiny, a modernized CNN architecture that incorporates depthwise convolutions with larger $7\times7$ kernels together with other design elements inspired by Vision Transformers~\cite{Liu_2022_CVPR}, achieved the highest MCC values across most training strategies. With all available real defects, pure oversampling achieved a maximum MCC of 0.98. EfficientNet-B4 trained with all available defects also reached an MCC of 0.98. Under these full-data conditions, the mixed training strategy achieved a comparable level of performance.

For all other architectures, mixed training with all available defects yielded higher MCC values, particularly in standard backbones such as ResNet~\cite{He.2016}, achieving 0.96 (ResNet-18) and 0.97 (ResNet-50), compared to 0.91 and 0.93 with pure oversampling.

In the few-shot regime (10 real defects), pure oversampling was outperformed by the mixed training strategy by large margins, with MCC improvements of 0.16 (ConvNeXt-Tiny) to 0.21 (ResNet-18). Across all backbones, starting from the purely synthetic baseline, the injection of real defects into the balanced synthetic stream progressively improved defect detection. Even a few injected real defects proved beneficial, achieving MCC values greater than 0.91 across all backbones. Moreover, the ConvNeXt-Tiny architecture reached a high MCC of 0.96 with just 10 injected real defects, highlighting the superiority of the mixed strategy in few-shot regimes.

Furthermore, the synthetic-only baseline, which did not rely on any real defects for defect generation (model-free approach), outperformed few-shot (10 real defects) oversampling in 4 out of 5 backbone experiments.

Based on the given industrial requirements and to avoid any inference bottleneck, the best accuracy–efficiency trade-off was achieved with the ResNet-18 architecture, reaching performance close to that of more modern architectures when using the mixed training strategy on all available data. Nonetheless, depending on future industrial requirements, ConvNeXt architectures remain a viable choice, delivering industry-applicable detection performance even in the few-shot regime.

\begin{table*}[h]
    \centering
    \caption{Defect detection performance of the investigated mixed supervised training strategy~\cite{HaselmannKrassnig.2022} evaluated on LSM-1. Starting from purely synthetic training, real defective patches were progressively injected with an injection probability $p_{\text{inj}}$ of 1/32 at varying fractions (1/16, 1/8, 1/4, 1/2, 1) of collected defects. The best-performing fraction, based on image-level MCC, is highlighted in bold.}
    \resizebox{0.8\textwidth}{!}{
        \begin{tabular}{ccccccccc}
            \toprule[1pt]
            \textbf{Defect Group} & \textbf{\#Real Defects} & \textbf{MCC}  & \textbf{Recall (\%)} & \textbf{FPR (\%)} & \textbf{TN} & \textbf{TP} & \textbf{FN}  & \textbf{FP}  \\
            \midrule[0.5pt]
            -                     & \phantom{0}\phantom{0}0 & 0.81          & 76.8                 & 0.7               & 1459        & 73          & 22           & 11           \\
            points                & \phantom{0}10           & 0.86          & 82.1                 & 0.5               & 1463        & 78          & 17           & \phantom{0}7 \\
            area                  & \phantom{0}10           & 0.90          & 84.2                 & 0.2               & 1467        & 80          & 15           & \phantom{0}3 \\
            mixed                 & \phantom{0}10           & 0.91          & 84.2                 & 0.1               & 1469        & 80          & 15           & \phantom{0}1 \\
            mixed                 & \phantom{0}21           & 0.91          & 86.3                 & 0.2               & 1467        & 82          & 13           & \phantom{0}3 \\
            mixed                 & \phantom{0}43           & 0.93          & 91.6                 & 0.3               & 1466        & 87          & \phantom{0}8 & \phantom{0}4 \\
            mixed                 & \phantom{0}86           & 0.93          & 90.5                 & 0.2               & 1467        & 86          & \phantom{0}9 & \phantom{0}3 \\
            mixed                 & 172                     & \textbf{0.96} & 95.8                 & 0.2               & 1467        & 91          & \phantom{0}4 & \phantom{0}3 \\
            \bottomrule[1pt]
        \end{tabular}}%
    \label{tab2}
\end{table*}%



\begin{table*}[h]
    \centering
    \caption{Defect detection performance of the investigated mixed supervised training strategy~\cite{HaselmannKrassnig.2022} evaluated on ASM. Starting from purely synthetic training, real defective patches were progressively injected with an injection probability $p_{\text{inj}}$ of 1/32 at varying fractions (1/16, 1/8, 1/4, 1/2, 1) of collected defects. The best-performing fraction, based on image-level MCC, is highlighted in bold.}
    \resizebox{0.8\textwidth}{!}{
        \begin{tabular}{ccccccccc}
            \toprule[1pt]
            \textbf{Defect Group} & \textbf{\#Real Defects} & \textbf{MCC}  & \textbf{Recall (\%)} & \textbf{FPR (\%)} & \textbf{TN} & \textbf{TP} & \textbf{FN}  & \textbf{FP}  \\
            \midrule[0.5pt]
            -                     & \phantom{0}\phantom{0}0 & 0.70          & 62.2                 & 0.6               & 1905        & 46          & 28           & 11           \\
            points                & \phantom{0}\phantom{0}9 & 0.76          & 66.2                 & 0.3               & 1910        & 49          & 25           & \phantom{0}6 \\
            area                  & \phantom{0}\phantom{0}9 & 0.91          & 91.9                 & 0.4               & 1909        & 68          & \phantom{0}6 & \phantom{0}7 \\
            mixed                 & \phantom{0}\phantom{0}9 & 0.94          & 93.2                 & 0.2               & 1912        & 69          & \phantom{0}5 & \phantom{0}4 \\
            mixed                 & \phantom{0}18           & 0.92          & 91.9                 & 0.3               & 1911        & 68          & \phantom{0}6 & \phantom{0}5 \\
            mixed                 & \phantom{0}36           & \textbf{0.97} & 95.9                 & 0.1               & 1914        & 71          & \phantom{0}3 & \phantom{0}2 \\
            mixed                 & \phantom{0}73           & 0.96          & 94.6                 & 0.1               & 1914        & 70          & \phantom{0}4 & \phantom{0}2 \\
            mixed                 & 146                     & 0.96          & 97.3                 & 0.2               & 1912        & 72          & \phantom{0}2 & \phantom{0}4 \\
            \bottomrule[1pt]
        \end{tabular}}%
    \label{tab3}
\end{table*}%

\subsubsection{Scalability and Synthetic Defect Complementation}\label{sec521}

Based on the ablation, ResNet-18 was selected for the subsequent experiments, as it represented the best accuracy-efficiency trade-off. The scalability of mixed training was evaluated on LSM-1, LSM-2, and ASM. Experiments examined whether small numbers of real defects with previously unseen characteristics—such as squeegee strokes in ASM or grid defects in LSM-1, which could not be synthesized—could complement synthetic features and improve generalization.

Training with purely synthetic defects in LSM-1 resulted in an MCC of 0.81 with an FPR of 0.7~\%. As described in the previous publication~\cite{Krassnig.2024}, depending on the underlying preprocessing strategy, FPRs below 0.59~\% and 0.33~\%, respectively, were necessary to limit unwanted false alarms, leading to unnecessary sample rejections.

Gradually adding real defective patches increased performance, achieving an MCC of 0.96 when utilizing all 172 available defects. This reduced the number of FN predictions from 22 to 4, representing an 81~\% reduction. Using only 10 defective patches from both defect groups (mixed) resulted in an MCC of 0.91. For comparison, supervised training with oversampling on all 172 available real defects achieved the same MCC of 0.91 (Table~\ref{tab:ablation}).

Training with separated fractions of point and area defects also showed improved detection performance, with area defects having more impact. Adding this defect group reduced group-related FNs from 6 in purely synthetic training to 1 in both mixed training and area-specific fractions. In mixed fractions greater than 1/8, no area defect was overlooked.

Training on the LSM-2 modality achieved near-perfect detection performance, overlooking only a single defective patch. The injection of real defective patches accelerated convergence, requiring only 10 epochs compared to approximately 20 epochs for purely synthetic training. This behavior may be attributed to the clearly pronounced defect characteristics of pinholes and pattern misalignment, as illustrated in Fig.~\ref{fig1} and Fig.~\ref{fig2}.

The ASM modality can be defined as the most demanding dataset due to its specific illumination characteristics (see Fig.~\ref{fig2}) and imperfect segmentation masks (see Fig.~\ref{fig_asm_artifacts}). Training on purely synthetic defects resulted in an MCC of 0.70, which could be significantly improved to greater than 0.95 by utilizing fractions upwards of 1/8. This improvement reduced FNs by up to 93~\% through the injection of real defects. The FPRs were comparable to those observed in LSM-1, indicating robustness to permitted sample variations and challenging inspection conditions.

Leveraging the area defect group or mixed fractions had a significantly greater impact on detection performance compared to relying solely on punctual defects. In purely synthetic training, 23 out of 28 FNs were area defects. Training with injected area defects reduced this number to only a few (up to 3). Utilizing all 146 available defects resulted in just 2 FNs, one of each defect group.

Additional experiments using solely real defects within the supervised oversampling approach~\cite{Krassnig.2024} achieved MCCs greater than 0.98 for fractions upwards of 1/4. The more distinct defect characteristics compared to LSM-1 (e.g. scratches or dots), as well as the limited defect diversity within the test set (see Section~\ref{sec325}), might be reasonable explanations for observed performance saturation.

As observed across other backbones in LSM-1 (Table~\ref{tab:ablation}), the performance of the proposed mixed training strategy remained superior in the low data regimes (fractions 1/16 and 1/8), especially when training was performed exclusively on area and point defect groups. In these settings, MCCs of 0.91 and 0.76 were achieved, respectively (see Table~\ref{tab3}), compared to MCCs of 0.56 and 0.65 for pure oversampling. Furthermore, optimized F1 thresholds below 0.1 indicated majority-class bias in the supervised oversampling strategy in these few-shot regimes.

\paragraph{Complementation with Real Defects}

As described earlier, the capability to synthesize area defects is limited. In the case of ASM, defect synthesis for the squeegee stroke defect class (Fig.~\ref{fig2}) was not possible and can be characterized by having unseen feature distributions. In contrast, LSM-1 contains grid defects (Fig.~\ref{fig2}) that exhibit similar punctual features to the synthesized ones. Models trained with synthetic defects generalized better on LSM-1, whereas ASM benefited more strongly from the injection of area defects.

For the ASM modality, the injection of 9 area defects led to an increase in the MCC by 0.21, reaching 0.91. The injection of similar synthesized feature distributions, as in the case of point defects, proved beneficial for the overall defect detection performance, including previously overlooked punctual defects. Furthermore, training on several tens of thousands of fault-free patches introduced permitted normal feature variability, resulting in industry-applicable FPRs of less than 0.59 respectively 0.33~\%.

\subsubsection{Key Findings Mixed Training Strategy}

Above experiments indicate that mixed training achieves scalable performance across all investigated modalities, with synthetic and real defects complementing each other.
(i) In LSM-1, model-free synthesis established a strong initial boundary, achieving an MCC of up to 0.92 in the synthetic-only baseline with ConvNeXt-Tiny. Its capability, however, appeared limited by the synthesis method and the underlying modality: in LSM-1, punctual defect characteristics such as those occurring in grid defects could be captured to a significant degree synthetically, whereas in ASM, unsynthesizable morphologies such as squeegee strokes required real injection.
(ii) Injecting even a few real defects complemented the already learned synthetic features, reducing false negatives and refining the initial boundary.
(iii) Mixed training consistently showed superior performance in few-shot regimes, achieving an MCC of at least 0.91 with only 10 real defect samples across various backbones when evaluated on LSM-1. It was beneficial in avoiding majority-class bias, converged faster, and maintained the low FPRs required in industrial inspection. The mixed training strategy proved architecture-agnostic, improving results across ResNet~\cite{He.2016}, EfficientNet~\cite{Eff.2019}, and ConvNeXt~\cite{Liu_2022_CVPR}, with clear trade-offs between detection performance and inference speed.

Together, these findings suggest that synthetic defects can serve as a cold-start baseline, while injected real defects provide complementary information for defect characteristics that cannot be synthesized. Furthermore, these results demonstrate that synthetic data can substantially reduce the number of real samples needed to achieve a given level of detection performance.

Thus, model-free synthesis approaches, as implemented in the proposed algorithm~\cite{Haselmann.2017}, which does not require defective training data for elaborate model training, may represent a viable option for rapid deployment in industry, as demonstrated in the proposed industrial use case~\cite{Krassnig.2024}.

\subsubsection{Limitations and Future Adaptations}\label{sec522}
The above results are based on the initial random selection of defective samples and are therefore expected to vary with different seeds during dataset generation. Extensive hyperparameter tuning and statistical analysis, including multi-seed training, are left for future work. In addition, the distribution of available ``area'' and ``points'' defect groups is imbalanced, due to limited availability (Table~\ref{tabA1}).

While this approach avoids the need for a trained generative model, hyperparameter tuning of the defect synthesis algorithm can be elaborate, and the diversity of synthesized defects remains limited.

Depending on data availability and deployment stage, more recent approaches such as generative models or vision language models~\cite{xu2025survey} could be integrated as synthesis modules within this adaptable closed-set strategy. A combination of model-free approaches (e.g. based on Perlin masks~\cite{Perlin.1985}) with model-based approaches (e.g. diffusion-based models~\cite{Gui.2025}) could further enhance feature diversity. ISP-AD therefore provides a benchmark, with its additional fault-free data splits enabling the investigation of emerging synthesis methods (see Appendix~\ref{append}).

In addition, future work will include quantitative comparisons of stochastic batch-level injection with dataset-level or deterministic batch mixing strategies~\cite{Dey.2024,huang2025weakly}. While our approach differs from continual or class-incremental learning~\cite{BALZATEGUI2023120382}, these strategies will also be explored to address related challenges in efficient and adaptive training. Furthermore, applying the injection scheme to additional industrial datasets and manufacturing domains may offer further insights into its general applicability.

\subsection{Unsupervised Methods}\label{sec53}

To estimate the defect detection performance of unsupervised methods in comparison to the supervised approach described above, evaluations were conducted using the same imbalanced test datasets. In addition to image-level labels, ground truth masks were utilized to assess the segmentation performance of each method. Performance metrics of various SOTA unsupervised approaches, applied to modalities: LSM-1, LSM-2 and ASM, are summarized in Tables~\ref{tab4} to~\ref{tab6}.

\subsubsection{Results}\label{sec531}

\begin{table*}[t]
    \centering
    \caption{Defect detection performance of investigated unsupervised approaches (EfficientAD~\cite{Batzner.2024}, U-Flow~\cite{Tailanian.2024}, GLASS~\cite{Chen.2024}, and DDAD~\cite{Mousakhan.25.05.2023}) on LSM-1. The best-performing method, based on image-level MCC, is highlighted in bold.}
    \resizebox{0.9\textwidth}{!}{
        \begin{tabular}{lccccccc}
            \toprule[1pt]
            \textbf{Method} & \textbf{AUROC (\%)}             & \textbf{MCC}                    & \textbf{Recall (\%)} & \textbf{Precision (\%)} & \textbf{FPR (\%)} & \textbf{AUROC (\%)} & \textbf{PRO (\%)} \\
            \midrule[0.5pt]
                            & \multicolumn{5}{c}{image-level} & \multicolumn{2}{c}{pixel-level}                                                                                                                \\
            \cmidrule[0.5pt](lr){2-6}
            \cmidrule[0.5pt](lr){7-8}
            EfficientAD     & 92.6                            & 0.87                            & 82.1                 & 94.0                    & 0.3               & 80.8                & \phantom{0}4.1    \\
            U-Flow          & 95.9                            & 0.77                            & 75.8                 & 80.9                    & 1.2               & 99.1                & 28.3              \\

            GLASS           & 99.5                            & \textbf{0.89}                   & 92.6                 & 86.3                    & 1.0               & 99.2                & 13.0              \\
            DDAD            & 91.9                            & 0.60                            & 63.2                 & 62.5                    & 2.4               & 93.7                & 16.8              \\
            \bottomrule[1pt]
        \end{tabular}}%
    \label{tab4}
\end{table*}%

\begin{table*}[t]
    \centering
    \caption{Defect detection performance of investigated unsupervised approaches (EfficientAD~\cite{Batzner.2024}, U-Flow~\cite{Tailanian.2024}, GLASS~\cite{Chen.2024}, and DDAD~\cite{Mousakhan.25.05.2023}) on LSM-2. The best-performing method, based on image-level MCC, is highlighted in bold.}
    \resizebox{0.9\textwidth}{!}{
        \begin{tabular}{lccccccc}
            \toprule[1pt]
            \textbf{Method} & \textbf{AUROC (\%)}             & \textbf{MCC}                    & \textbf{Recall (\%)} & \textbf{Precision (\%)} & \textbf{FPR (\%)} & \textbf{AUROC (\%)} & \textbf{PRO (\%)} \\
            \midrule[0.5pt]
                            & \multicolumn{5}{c}{image-level} & \multicolumn{2}{c}{pixel-level}                                                                                                                \\
            \cmidrule[0.5pt](lr){2-6}
            \cmidrule[0.5pt](lr){7-8}
            EfficientAD     & 99.1                            & \textbf{0.96}                   & 94.8                 & 98.9                    & 0.3               & 80.9                & 58.8              \\
            U-Flow          & 99.9                            & \textbf{0.96}                   & 95.8                 & 97.9                    & 0.5               & 99.8                & 95.6              \\
            GLASS           & 99.5                            & 0.94                            & 96.9                 & 93.0                    & 1.8               & 99.9                & 99.2              \\
            DDAD            & 98.2                            & 0.82                            & 84.4                 & 87.1                    & 3.1               & 99.7                & 77.9              \\
            \bottomrule[1pt]
        \end{tabular}}%
    \label{tab5}
\end{table*}%

\begin{table*}[t]
    \centering
    \caption{Defect detection performance of investigated unsupervised approaches (EfficientAD~\cite{Batzner.2024}, U-Flow~\cite{Tailanian.2024}, GLASS~\cite{Chen.2024}, and DDAD~\cite{Mousakhan.25.05.2023}) on ASM. The best-performing method, based on image-level MCC, is highlighted in bold.}
    \resizebox{0.9\textwidth}{!}{
        \begin{tabular}{lccccccc}
            \toprule[1pt]
            \textbf{Method} & \textbf{AUROC (\%)}             & \textbf{MCC}                    & \textbf{Recall (\%)} & \textbf{Precision (\%)} & \textbf{FPR (\%)} & \textbf{AUROC (\%)} & \textbf{PRO (\%)} \\
            \midrule[0.5pt]
                            & \multicolumn{5}{c}{image-level} & \multicolumn{2}{c}{pixel-level}                                                                                                                \\
            \cmidrule[0.5pt](lr){2-6}
            \cmidrule[0.5pt](lr){7-8}
            EfficientAD     & 85.5                            & 0.31                            & 28.4                 & 38.9                    & 1.7               & 88.4                & \phantom{0}4.5    \\
            U-Flow          & 92.4                            & \textbf{0.49}                   & 52.7                 & 48.8                    & 2.1               & 89.3                & 19.9              \\
            GLASS           & 87.4                            & 0.34                            & 48.6                 & 28.6                    & 4.7               & 82.6                & \phantom{0}4.0    \\
            DDAD            & 90.9                            & 0.38                            & 55.4                 & 29.7                    & 5.1               & 96.6                & \phantom{0}6.5    \\
            \bottomrule[1pt]
        \end{tabular}}%
    \label{tab6}
\end{table*}%

Image-level defect detection performance in LSM-1 across different methods shows substantial variation with MCCs ranging from 0.60 at an FPR of 2.4~\% to 0.89 at an FPR of 1.0~\%. At the pixel level, minimum PRO scores ranged from 4.1 to 28.3~\%. In general, the low PRO-scores highlight the challenging task of detecting small punctual anomalies such as grid defects in Fig.~\ref{fig2}. Common FPs occur at masked border regions or within areas of permitted variations in patterned structures (Fig.~\ref{fig4}).
\begin{figure*}[h]\label{fig4}
    \centering
    \includegraphics[width=1.0\textwidth]{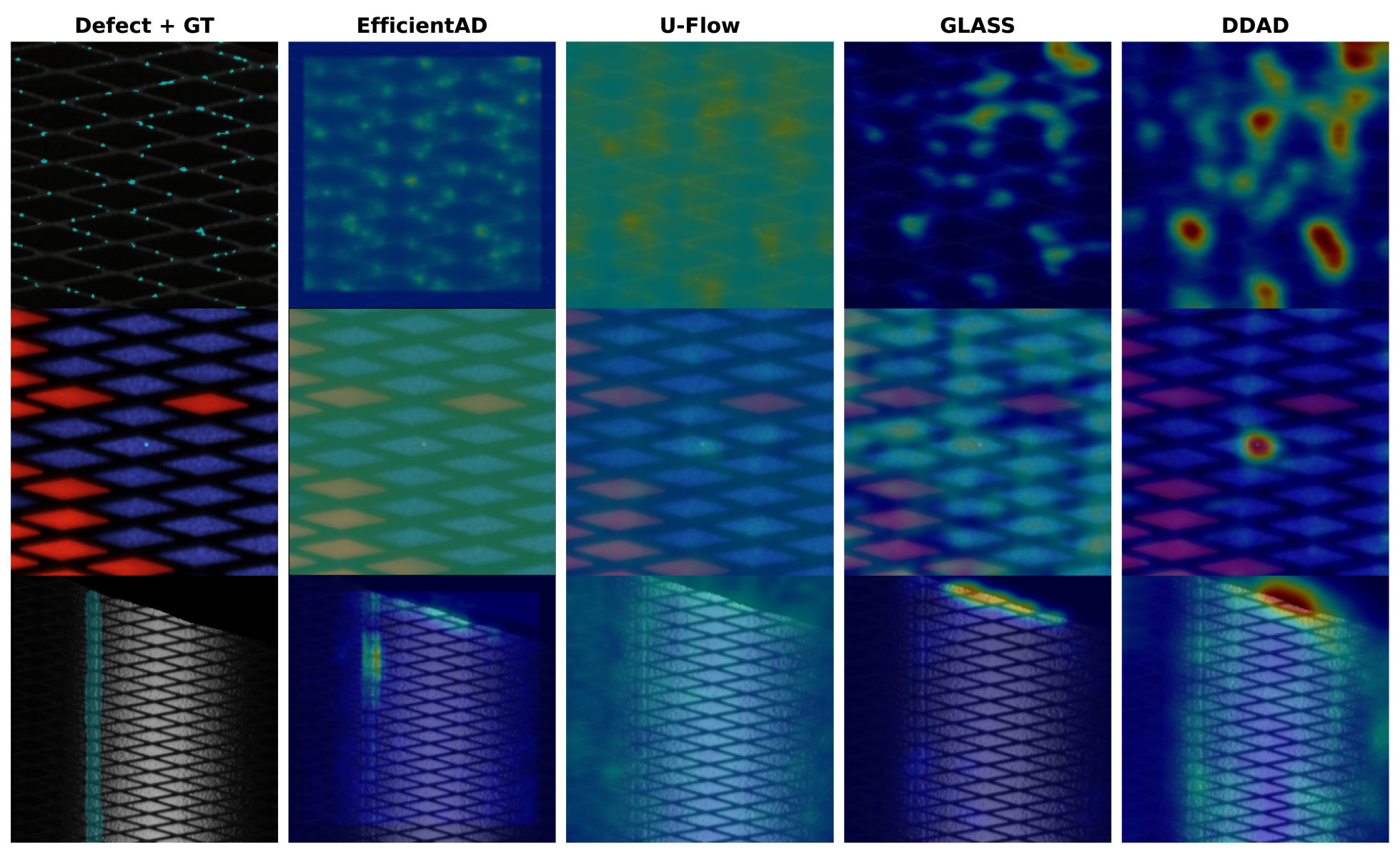}
    \caption{Illustration of the pixel-level detection performance (anomaly map overlay) for the evaluated SOTA unsupervised methods (EfficientAD~\cite{Batzner.2024}, U-Flow~\cite{Tailanian.2024}, GLASS~\cite{Chen.2024}, and DDAD~\cite{Mousakhan.25.05.2023}) on LSM-1, LSM-2, and ASM (top to bottom). The first column displays defects with their ground truth overlays (grid, pinhole, squeegee stroke). Overestimated defect areas (e.g., grid defect) and missed detections of small, low-contrast pinholes demonstrate the challenges of proposed industrial scenario. False positives frequently occur in masked border regions and areas of high contrast variations, as observed in ASM (bottom row).}
    \label{fig4}
\end{figure*}
The comparison of high pixel-level AUROCs ($>$ 99~\%) compared to the measured PRO-scores illustrates their misleading interpretability in imbalanced scenarios.

EfficientAD demonstrated a 7-fold reduction of FPR, compared to the reconstruction based approach (DDAD) based on denoising diffusion. Thus, EfficientAD demonstrated an industry-applicable FPR of 0.3~\%, with qualitatively precise anomaly segmentations of small pronounced defects. However, despite the patch descriptor's well-defined receptive field ensuring robustness, small defects remain boundary cases and are prone to misclassification. This is likely due to the bias introduced by ImageNet pretraining and the shallow patch descriptor architecture, which may limit its ability to extract fine-grained, in-domain features for subtle anomalies.
In general, small and weakly contrasted point defects remain challenging even in GLASS, resulting in a recall of 92.6~\%.

Compared to modality LSM-1, defect classes such as pinholes and pattern misalignments in LSM-2 (Fig.~\ref{fig1} and Fig.~\ref{fig2}), which stand out prominently from the background pattern, exhibit more pronounced defect features. Consequently, both image- and pixel-level performance of various unsupervised approaches show improvements, achieving MCCs of up to 0.96 and PRO scores above 99~\%.

However, segmentation performance varies significantly across the underlying approaches. Small punctual features are often overestimated or even missed, particularly in embedding-based methods. DDAD demonstrates qualitatively enhanced reconstruction capability, however, it struggles with patches that exhibit large background variations. Although indicating solid segmentation performance, PRO-scores do not consider FP assignments, thus overestimations of defective regions are not directly penalized.

ASM consists of numerous patches containing imperfect masked image borders (Fig.~\ref{fig_asm_artifacts}), due to varying imaging conditions and the subsequent influence on segmentation performance of the applied deep learning-based model. These artifacts reflect practical deployment conditions and were intentionally retained in the dataset to simulate realistic segmentation imperfections. Additionally, this modality is characterized by large normal feature variations caused by the underlying imaging procedure, which relies on direct reflection.

All approaches exhibited misclassifications in these regions, often leading to TP assignments due to coincidences of real defective areas within the same patch. As a result, despite achieving low performance among all modalities, overestimations of image-level metrics must be considered. In addition to small punctual defects, it was particularly difficult to detect pronounced squeegee strokes among the methods evaluated. However, the anomaly maps (Fig.~\ref{fig5}) indicate that the student-teacher based approach EfficientAD achieved the best qualitative segmentation performance.
Again, the localized patch descriptor seems preferable for suppressing false positives.

Above conducted experiments revealed that the defect detection performance of leveraged unsupervised approaches is strongly dependent on dataset properties, determined by its underlying imaging modalities, preprocessing and defect appearance.

Showing distinct defective feature characteristics, LSM-2 achieved the best overall performance, resulting in industrially applicable FPRs for EfficientAD and U-Flow, although struggling with weakly contrasted pinholes. Compared to the supervised approach in Section~\ref{sec41}, FPRs are mostly significantly higher (up to factor 10 in ASM), limiting industrial applicability, particularly in the case of ASM.

Despite DDAD's qualitatively good reconstruction capability, modelling permitted pattern variations and patterns in border regions was often not feasible (Fig.~\ref{fig6}). In addition, DDAD exhibited reconstruction instabilities resulting in noise that increased with higher domain adaptation weightings. These effects may be attributable to the loss of high-frequency detail induced by receptive-field limitations, as well as to characteristics of the denoising parametrization.

\subsubsection{Limitations}\label{sec532}
The evaluations of these approaches should be considered as estimations based on the chosen methods parameters, preprocessing techniques, and training settings, including random seeds. Extensive hyperparameter tuning and advanced preprocessing strategies are left for future research.

\section{Conclusions and Outlook}\label{sec6}

This publication introduces a novel surface defect detection dataset, ISP-AD, which represents a real-world industrial use case. The investigated screen-printed samples were captured using three different optical modalities. Typical for industrial inspection, the dataset reflects imperfect imaging conditions caused by imaging modalities, data preprocessing, and permitted sample variations, thereby presenting challenges for defect detection algorithms. In addition to extending common unsupervised settings, the dataset includes large-scale supervised training data, comprising tens of thousands of fault-free patches. Based on a review of recent literature, the proposed dataset can be assumed to be the largest publicly available, enabling research on both unsupervised and supervised approaches by using both synthetic and real defects.

It was observed that SOTA unsupervised methods, despite achieving remarkable performance on established benchmarks such as MVTec~\cite{PaulBergmann.} (image-level AUROC $>$ 99~\%), struggle on ISP-AD, with modality-specific image-level AUROC dropping to 85.5~\% on ASM and pixel-level PRO scores falling to 4.0 -- 4.5~\% (on the imbalanced test set; see Table~\ref{tabA1}). The occurrence of small and weakly contrasted defects, combined with high permitted design variability, significantly limited defect detection performance, especially at the pixel level.

Among the evaluated methods, the most promising approaches are based on defect synthesis at both the feature and image-levels and knowledge distillation using patch descriptor networks, achieving image-level MCCs of up to 0.89 in LSM-1 and 0.96 in LSM-2. However, despite robust performance of EfficientAD~\cite{Batzner.2024} with industry-applicable low FPRs in LSM-1 and LSM-2, weakly contrasted defects remained challenging. While certain unsupervised approaches have shown promising advances in image-level detection performance for specific imaging modalities (e.g., LSM-2), future research should focus enhancing segmentation performance under industrial imaging conditions.

Incorporating available weakly labeled defective samples within a mixed supervised training strategy is able to improve image-level defect detection performance by large margins. The experiments described in Section~\ref{sec52} indicate that the initial comprehensive feature distribution, by means of several tens of thousands of synthesized punctual defects, is complemented and made more diverse, even with small amounts of available real defects. The injection of small fractions (9 samples in ASM) of previously unseen defect classes (area defects) substantially improved model generalization. These results indicate that previously unseen defects, such as those that could not be synthesized, can provide strong learning signals, helping to refine the decision boundary.

The defect detection performance of small and weakly contrasted point defects was substantially improved with mixed training (MCC = 0.96 in LSM-1), compared to training solely on real defects (MCC = 0.91) using a standard ResNet-18 backbone. Notably, in the few-shot regime, the mixed supervised strategy showed substantial performance gains compared to pure oversampling across various backbones (ResNet-18, ResNet-50, EfficientNet-B0, EfficientNet-B4, and ConvNeXt-Tiny). More recent backbones, such as ConvNeXt-Tiny, achieved a high MCC of up to 0.96 with only 10 real injected defects.

These findings suggest a valuable capability for industrial applications, as emerging defects, both seen and previously unseen, can be efficiently integrated into subsequent scalable training, enhancing overall feature diversity. Thus, initial training on synthetic defects can be incrementally improved, meeting process requirements such as a low FPR and high recall.

Furthermore, purely synthetic training achieved near-optimal (MCCs up to  0.99) defect detection performance in modalities that exhibit distinct feature characteristics, such as pinholes in LSM-2. Purely synthetic training outperformed SOTA unsupervised approaches in two out of three modalities (LSM-2 and ASM; ResNet-18).

Despite promising investigations of generalization capabilities with a limited number of available defective samples, further experiments are necessary for deeper insights. These should involve more diverse defect classes and varying hyperparameters, such as the injection probability $p_{\text{inj}}$ of real defects. The selected injection probability of 1/32 was intended to avoid overfitting to injected real defect characteristics. However, in some cases—depending on defect availability and diversity, the defect synthesis approach, or the imaging modality (e.g., ASM)—higher injection probabilities may be preferable to align the decision boundary more closely with the incorporated defect features. Furthermore, in future work, defect selection strategies inspired by active learning frameworks~\cite{Gao.2025} could be considered, as higher fractions of real defects appear to saturate performance in certain cases. Therefore, injected diversity seems to be more important than the absolute quantity of defective data.

Additionally, permitted normal data variations are learned through large amounts (\num{50000} -- \num{100000} of augmented fault-free patches, even with noisy labels. Noisy labels, due to e.g. process contamination during the image acquisition, were estimated to account for 2 -- 5~\% in LSM-1. These corrupted labels did not show a significant impact during training. This observation is consistent with unsupervised and supervised learning paradigms: unsupervised methods rely on clean normal data to enable the modeling of compact representations, whereas supervised methods are able to learn a robust decision boundary even in the presence of noisy labels~\cite{bar2022spectral}. Regarding our approach, this can be explained by the large-scale balanced data streams, including a diverse set of permitted fault-free data variations, helping the model to learn what patterns are considered normal.

Thus, the screening effort of fault-free data in the proposed unsupervised training datasets could be considered more elaborate than the presented supervised data preprocessing workflow in~\cite{Krassnig.2024}, minimizing manual labeling effort to known defective sample regions.
Thus, further research should focus on efficient and industrially applicable data preparation and preprocessing workflows for unsupervised and supervised approaches, as well as on the influence of different proportions of noisy labels.
As a direct implication of above findings, research in self-supervised approaches that utilize these large-scale normal data distributions should be emphasized to enhance industry-applicable model robustness at low labeling costs.

In summary, the mixed supervised training strategy~\cite{HaselmannKrassnig.2022}, based on the efficient preprocessing workflow proposed in~\cite{Krassnig.2024}, is capable of fulfilling industrial inspection requirements with minimized labeling effort. Labeling is restricted to known defective samples and regions, as the extraction of fault-free patches is integrated into an automated procedure. The model-free synthesis algorithm enables defect generation without requiring elaborate learning of defect distributions. Complementary defect features are learned through the injection of a small set of weakly labeled real defects, enabling model generalization.
However, recent zero-shot and few-shot anomaly synthesis approaches~\cite{Shin.2025, Sun_2025_CVPR, lai2025anomalypainter} may be preferable in future adaptations, further enhancing defect realism and diversity within the synthetic baseline.

Based on our findings, we raise the question of whether defining industrial anomaly detection as a fully open-set problem is appropriate in common real-world manufacturing processes. Since defect causes are often known in advance, a reformulated paradigm such as the proposed restricted open-set problem (see Appendix~\ref{ROS-AD}) may offer a more practically applicable definition for industrial scenarios. Future work may explore this direction in more detail, particularly with regard to the formalization of such paradigms and the development of efficient learning and defect synthesis strategies considering realistic industrial constraints.

\subsection{Future Applications}\label{sec61}

The proposed dataset is designed to advance future research on both unsupervised and supervised approaches suitable for challenging industrial anomaly detection use cases. Leveraging the proposed dataset enables further exploration in the following directions:

\begin{itemize}
    \item Investigation of the impact of increased normal feature variations on the robustness of unsupervised approaches by utilizing larger sets of fault-free training samples than those used in the proposed experiments. Additionally, an examination of the effect of real noisy labels by applying the “unscreened” automatically extracted fault-free training data.
    \item Conduction of research on self-supervised learning approaches~\cite{Hojjati.2024} utilizing the synthetic or fault-free large-scale training splits. Therefore, feature extractors of unsupervised approaches, such as described in~\cite{Koshil.2024,Liang.2024}, could be pretrained to learn more discriminative features of the target domain. In particular, contrastive learning strategies, as introduced by~\cite{Liang.2024}, can exploit synthetic streams and their pixel-level ground-truth masks to enhance feature separation between normal and defective samples. Apart from their subsequent use in unsupervised approaches, these pre-trained extractors can be fine-tuned or employed in mixed training strategies, such as the proposed stochastic batch-level injection, to improve performance on downstream tasks.
    \item Foster research on zero-shot or few-shot defect synthesis approaches, such as recent works~\cite{zhang2024realnet,Shin.2025,Sun_2025_CVPR,lai2025anomalypainter}, for generating, e.g., large-scale synthetic streams applicable to self-supervised or mixed training strategies.
    \item Building on the proposed restricted open-set paradigm (see Appendix~\ref{ROS-AD}), knowledge of process-plausible defect characteristics (e.g., texture or shape) can serve as a conditioning constraint for defect-synthesis approaches~\cite{singh2025comprehensive}, including generative models. This may simplify defect synthesis by limiting the required diversity of defect patterns. Furthermore, training schemes based on cold-start synthetic training that subsequently transition into few-shot incremental learning strategies~\cite{zhang2025few} represent a viable alternative to the proposed mixed training strategy. Such strategies are consistent with the restricted open-set assumption and could facilitate more robust and adaptable industrial anomaly detection.
\end{itemize}

\backmatter





\bmhead{Acknowledgements}
The research work was performed within the COMET-project: Deep on-line learning for highly adaptable polymer surface inspection systems (project-no.: 879785) at the Polymer Competence Center Leoben GmbH (PCCL, Austria) within the framework of the COMET-program of the Federal Ministry for Climate Action, Environment, Energy, Mobility, Innovation and Technology and the Federal Ministry for Digital and Economic Affairs and with contributions by Burg Design GmbH. The PCCL is funded by the Austrian Government and the State Governments of Styria, Lower Austria and Upper Austria. The authors thank the current and former members of our research group for their support and helpful discussions.

\bmhead{Funding} Open access publishing enabled by Montanuniversität Leoben agreement with Springer Nature (Austria KEMÖ: Springer Transformative Agreement). This work was supported within the COMET-project (project-no.: 879785).

\bmhead{Data availability} \label{data} The dataset supporting this research is publicly available on Zenodo at \url{https://doi.org/10.5281/zenodo.14911042}, and is licensed under the Creative Commons Attribution-NonCommercial-ShareAlike 4.0 International (CC BY-NC-SA 4.0).

\bmhead{Author contribution}
Conceptualization: PJK and DPG; Methodology: PJK; Formal analysis and investigation: PJK; Writing - original draft preparation: PJK; Writing - review and editing: PJK and DPG; Funding acquisition: DPG; Supervision: DPG.

\section*{Declarations}

\bmhead{Competing interests} The authors have no competing interests to declare that are relevant to the content of this article.

\begin{appendices}

    \renewcommand{\thefigure}{\arabic{figure}}  
    \renewcommand{\thetable}{\arabic{table}}    

    \setcounter{table}{7}
    \setcounter{figure}{4}

    \section{}
    \label{append}

    Table~\ref{tabA1} illustrates the training and test data splits
    of the proposed ISP-AD dataset. The augmentation
    settings applied to the supervised training
    data are shown in Table~\ref{tabA2}. In addition to the
    presented data splits, large-scale fault-free training
    data (\num{10000} -- \num{20000} patches) without augmentation
    is made available for all three modalities:
    LSM-1, LSM-2, and ASM. As outlined in Section~\ref{sec32},
    the extracted patches can be considered noisy. The
    naming convention of image-level labels within the
    .hdf5 files follows:

    \begin{itemize}
        \item Stream including both fault-free and synthetic defects: \textbf{syn\_stream}
        \item Ground truth masks: \textbf{ground\_truth}
    \end{itemize}

    To reproduce the experiments presented in Section~\ref{sec52}, randomly sampled fractions of the mixed, area, and points defect groups will be made available separately in the dataset repository~(\nameref{data}).

    \begin{table*}[h]
        \centering
        \caption{Overview of the available data splits in the proposed ISP-AD dataset. The dataset is divided into unsupervised training data, containing screened fault-free patches, and supervised training data, which includes ``noisy'' fault-free patches along with additional synthetic and real defective data. Test splits remain the same for both settings, mimicking the imbalanced data distributions typical of industrial inspection. Additional unaugmented fault-free data available is not included in this listing.}
        \resizebox{0.9\textwidth}{!}{
            \begin{tabular}{ccccccccc}
                \toprule[1pt]
                                  & \multicolumn{5}{c}{\textbf{Train}}        & \multicolumn{3}{c}{\textbf{Test}}                                                                                                                                                           \\
                \cmidrule[0.5pt](lr){2-6}
                \cmidrule[0.5pt](lr){7-9}
                                  & \multicolumn{1}{l}{\textbf{Unsupervised}} & \multicolumn{4}{c}{\textbf{Supervised}} & \multicolumn{3}{c}{\textbf{Super- / Unsupervised}}                                                                                                \\
                \cmidrule[0.5pt](lr){2-2}
                \cmidrule[0.5pt](lr){3-6}
                \cmidrule[0.5pt](lr){7-9}
                \textbf{Modality} & \textbf{\#Good}                           & \textbf{\#Good}                         & \textbf{\#Synthetic}                               & \textbf{\#Area} & \textbf{\#Points} & \textbf{\#Good} & \textbf{\#Area } & \textbf{\#Points} \\
                \midrule[0.5pt]
                LSM-1             & 3678                                      & \num{100669}                            & \num{99331}                                        & 120             & \phantom{0}52     & 1470            & 49               & 46                \\
                LSM-2             & \phantom{0}401                            & \phantom{0}\num{50036}                  & \num{49964}                                        & \phantom{0}62   & \phantom{0}66     & \phantom{0}382  & 63               & 33                \\
                ASM               & \phantom{0}500                            & \num{103622}                            & \num{96369}                                        & \phantom{0}37   & 109               & 1916            & 33               & 41                \\
                \bottomrule[1pt]
            \end{tabular}}%
        \label{tabA1}%
    \end{table*}%

    \begin{table*}[h]
        \centering
        \caption{Augmentation settings for the generated supervised training sets, including random affine and illumination transformations.}
        \resizebox{0.9\textwidth}{!}{
            \begin{tabular}{cccccccc}
                \toprule[1pt]
                                  & \multicolumn{5}{c}{\textbf{Affine Transformations}} & \multicolumn{2}{c}{\textbf{Illumination Tranformations}}                                                                                                                \\
                \cmidrule[0.5pt](lr){2-6}
                \cmidrule[0.5pt](lr){7-8}
                \textbf{Modality} & \textbf{Rotation}                                   & \textbf{Scale}                                           & \textbf{Shear} & \textbf{Vertical Flip} & \textbf{Horizontal Flip} & \textbf{Brightness} & \textbf{Contrast} \\
                \midrule[0.5pt]
                LSM-1             & 45.0                                                & (0.90,1.10)                                              & 10.0           & yes                    & yes                      & (0.75,1.25)         & (0.75,1.25)       \\
                LSM-2             & 45.0                                                & (0.90,1.10)                                              & 20.0           & yes                    & yes                      & (0.75,1.25)         & (0.75,1.25)       \\
                ASM               & 10.0                                                & (1.0,1.2)                                                & 10.0           & yes                    & yes                      & no                  & no                \\
                \bottomrule[1pt]
            \end{tabular}}%
        \label{tabA2}%
    \end{table*}%

    \begin{figure*}[h]
        \centering
        \includegraphics[width=0.90\textwidth]{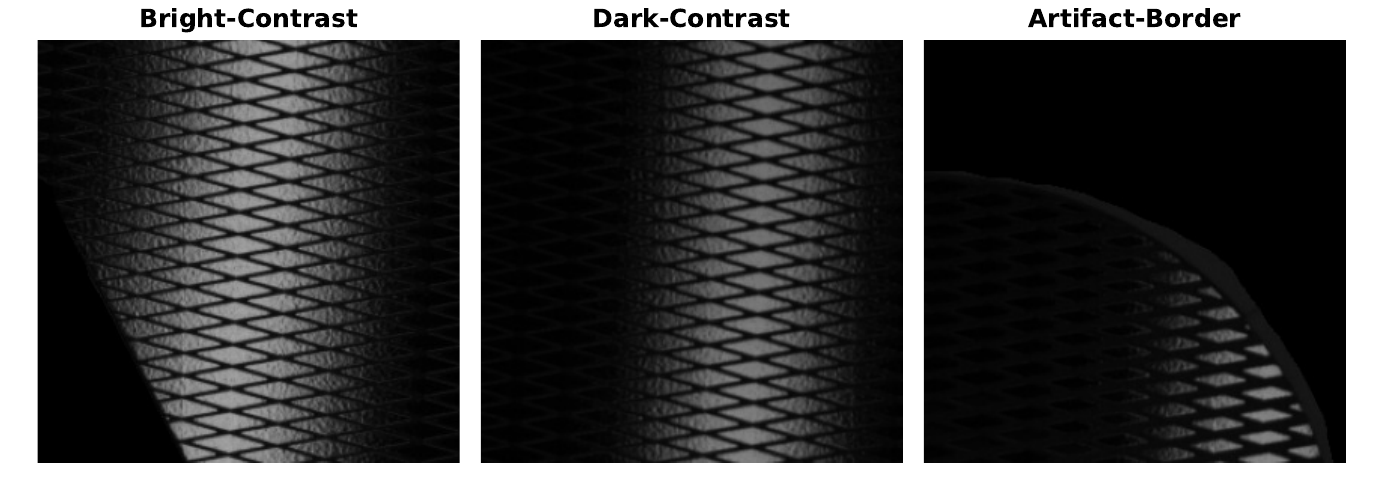}
        \caption{Depicted fault-free patches of the ASM modality illustrate challenging inspection scenarios in ISP-AD. Due to the illumination characteristics (brightfield imaging and sample positioning), textures exhibit strong contrast and intensity variations. Additionally, they are affected by permitted design variations inherent to the underlying screen printing process. The right patch shows irregularly masked border regions (segmentation artifacts), introduced by these varying conditions, which further contribute to feature variations.}
        \label{fig_asm_artifacts}
    \end{figure*}

    \begin{figure*}[h]
        \centering
        \includegraphics[width=0.95\textwidth]{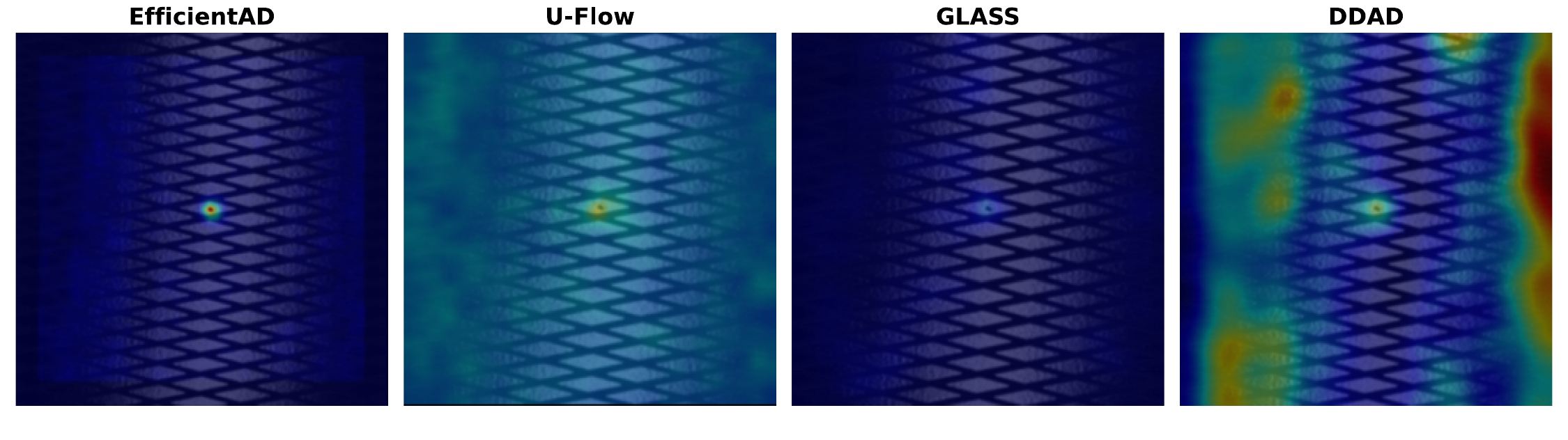}
        \caption{Visualization of the pixel-level detection performance (anomaly map overlay) for the investigated unsupervised methods (EfficientAD~\cite{Batzner.2024}, U-Flow~\cite{Tailanian.2024}, GLASS~\cite{Chen.2024}, and DDAD~\cite{Mousakhan.25.05.2023}) on ASM. EfficientAD demonstrated the most accurate detection of the depicted punctual defect. The reconstruction-based approach DDAD produced a high number of false positive detections in transmission areas of direct reflection and masked border regions.}
        \label{fig5}
    \end{figure*}

    \begin{figure*}[h]
        \centering
        \includegraphics[width=0.90\textwidth]{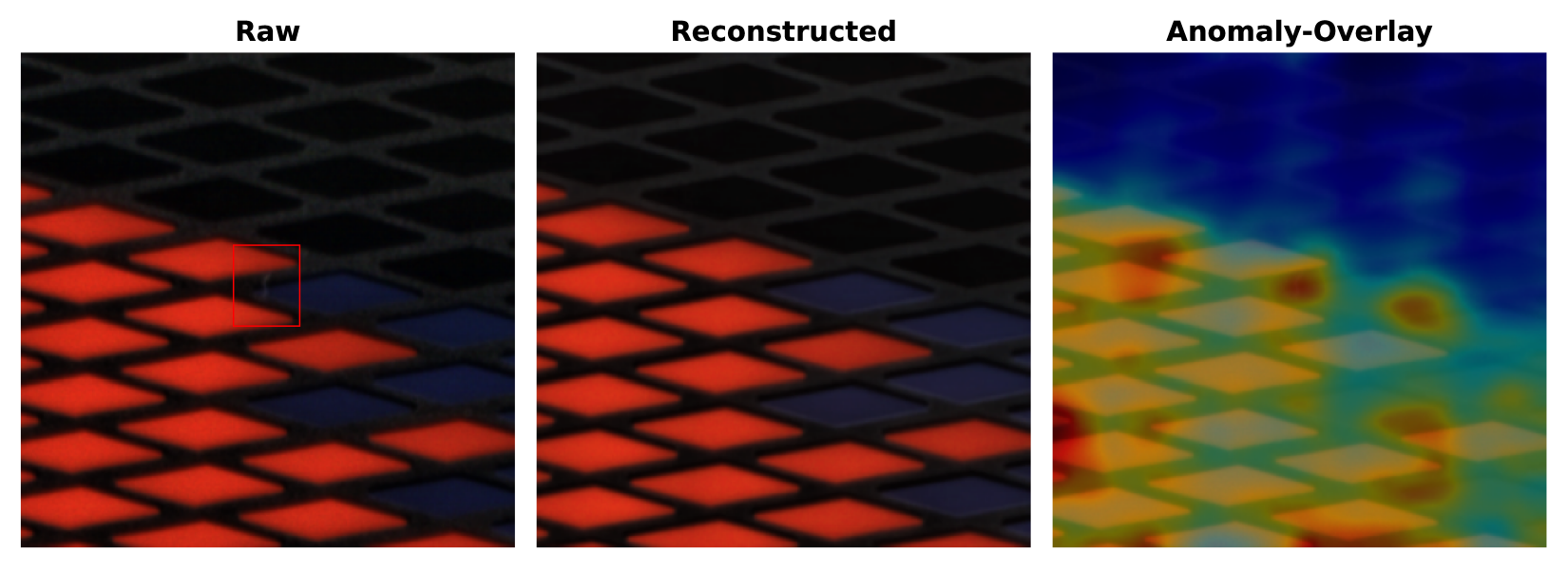}
        \caption{Illustration of the reconstruction capabilities of DDAD~\cite{Mousakhan.25.05.2023} on a patterned surface of LSM-1. The denoising diffusion process enabled a qualitatively good reconstruction, effectively omitting the small fiber in the raw image. However, fine-grained details, such as contrast variations in the grid regions, remained challenging, resulting in high anomaly scores across broad areas.
        }
        \label{fig6}
    \end{figure*}

    \subsection{The Restricted Open-Set Paradigm}
    \label{ROS-AD}

    Following~\cite{Ruff2021}, the normal data distribution \( P^{+} \) is concentrated in a bounded region \( \mathcal{N} \) of the data space \( \mathcal{X} \), referred to as the normal data space (concentration assumption~\cite{steinwart2005classification}). Despite being bounded, \( \mathcal{N} \) typically exhibits permitted variability due to process tolerances, and thresholds within this region may shift in response to changing quality requirements.

    Anomalies are assumed to be non-concentrated. A common open-set simplification assumes an uninformative anomaly prior~\cite{steinwart2005classification,Ruff2021}, modeling the anomaly distribution \( P^{-} \) as uniform over the (bounded) data space \( \mathcal{X} \).
    In modern industrial manufacturing processes, however, defects are not arbitrary but process-related.

    Previous work has mostly reflected this assumption only indirectly: datasets such as MVTec~\cite{PaulBergmann.} include exclusively process-related defects, and~\cite{Ruff2021,steinwart2005classification} point out the potential of more informed anomaly priors without explicitly formalizing a bounded anomaly domain.

    We therefore restrict the defect space to process-related characteristics, defined by
    \begin{equation}
        \mathcal{D} \;\subseteq\; \mathcal{X} \setminus \mathcal{N}.
        \label{eq:defect-domain}
    \end{equation}
    Out-of-domain defect characteristics are explicitly excluded.

    This formulation retains unseen but process-plausible defects while excluding unrelated anomalies. It does not assume uniformity within \( \mathcal{D} \). Furthermore, the entire possible defect space \( \mathcal{D} \) is initially unknown, in contrast to a closed-set classification paradigm.

    It therefore departs from the standard closed-set and open-set paradigms~\cite{Scheirer} and reflects realistic industrial inspection scenarios, where not all defect classes may be known in advance but are restricted to process-plausible characteristics.

    The investigated mixed training strategy (stochastic batch-level injection; see Section~\ref{sec41}) serves as an incremental mechanism to update the decision boundary within the restricted open-set data space \(\mathcal{N}\;\cup\; \mathcal{D} \), which comprises normal samples and process-related defects.

\begin{figure*}[h]
        \centering
        \includegraphics[width=0.90\textwidth]{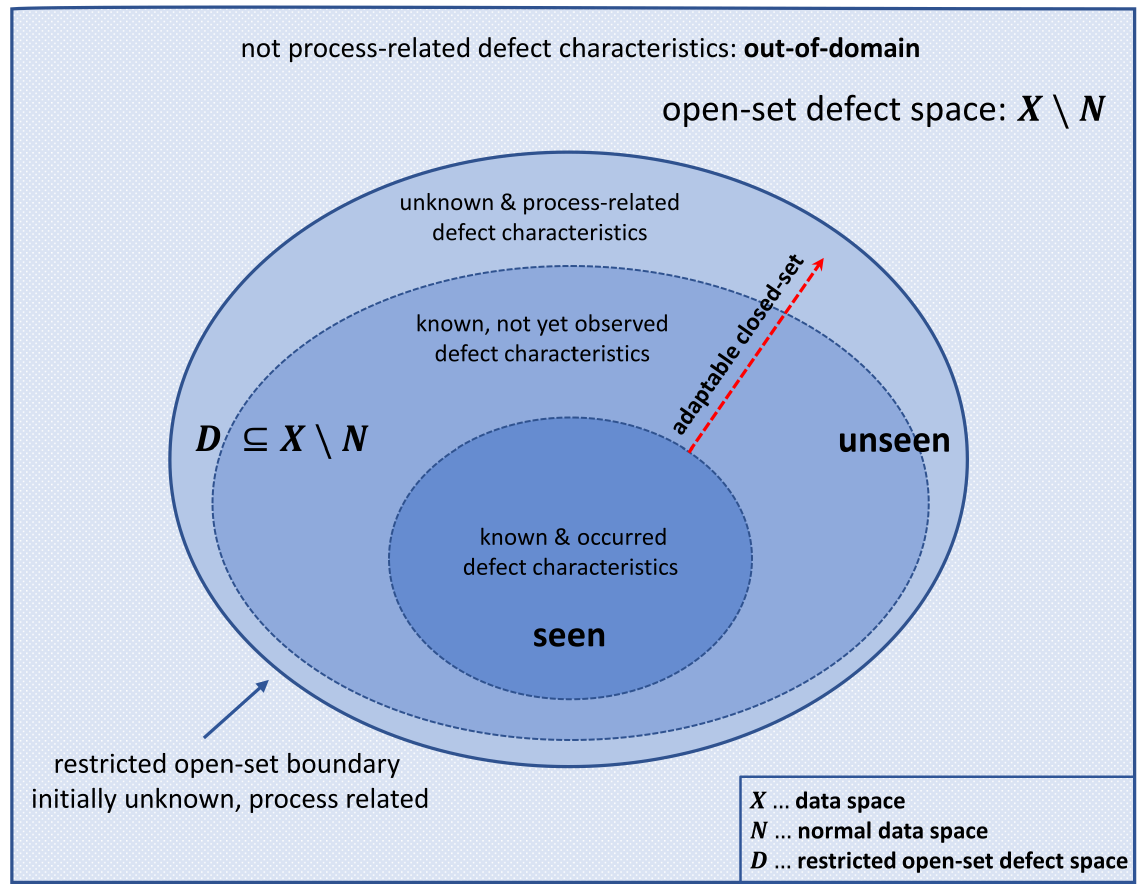}
        \caption{Contrast between the defect spaces in the proposed restricted open-set and the standard open-set anomaly-detection paradigms (see notation in Section~\ref{ROS-AD}). In the restricted open-set paradigm, the defect space \( \mathcal{D} \subseteq \mathcal{X} \setminus \mathcal{N} \) is limited to process-related defect characteristics. These defects may be (i) known and available for defect modeling or training, or (ii) previously unseen but still related to the underlying process. As unseen defects become available, they can be incorporated into the mixed training strategy using the stochastic batch-level injection scheme, thereby enlarging the set of seen defects within \(\mathcal{D}\). This training procedure can thus be referred to as an adaptable closed-set strategy, as it iteratively refines the already-learned decision boundary.}
        \label{fig7}
    \end{figure*}

    A more detailed formalization is left for future work, although the above notation may serve as a starting point. A schematic illustration of the restricted open-set paradigm is shown in Fig.~\ref{fig7}.

\end{appendices}


\clearpage

\end{document}